\documentclass[nohyperref]{article}
\usepackage{hyperref}

\usepackage[accepted]{icml2023}
\usepackage{microtype}
\usepackage{graphicx}
\usepackage{subfigure}
\usepackage{booktabs} 
\usepackage{multicol,multirow}
\usepackage{bbding}
\usepackage{amsmath}
\usepackage{amssymb}
\usepackage{mathtools}
\usepackage{amsthm}
\usepackage{tcolorbox}
\definecolor{dkblue}{rgb}{0,0.08,0.45}

\newcommand{\blue}[1]{\textcolor{dkblue}{#1}}

\theoremstyle{plain}
\newtheorem{theorem}{Theorem}[section]
\newtheorem{proposition}[theorem]{Proposition}

\theoremstyle{definition}

\newtheorem{assumption}[theorem]{Assumption}
\theoremstyle{remark}

\DeclareMathOperator*{\argmax}{arg\,max}
\newcommand{\meanN}{\frac{1}{n} \sum_{i=1}^n}
\newcommand{\sumA}{\sum_{a \in \mathcal{A}}}
\newcommand{\sumE}{\sum_{e \in \mathcal{E}}}
\newcommand{\sumC}{\sum_{c \in \mathcal{C}}}
\newcommand{\mE}{\mathbb{E}}
\newcommand{\mV}{\mathbb{V}}

\newcommand{\calD}{\mathcal{D}}
\newcommand{\calX}{\mathcal{X}}
\newcommand{\calA}{\mathcal{A}}
\newcommand{\calU}{\mathcal{U}_0}
\newcommand{\calC}{\mathcal{C}}
\newcommand{\calE}{\mathcal{E}}

\newcommand{\trueV}{V(\pi)}
\newcommand{\ips}{\hat{V}_{\mathrm{IPS}} (\pi; \calD)}
\newcommand{\mips}{\hat{V}_{\mathrm{MIPS}} (\pi; \calD)}
\newcommand{\offcem}{\hat{V}_{\mathrm{OffCEM}} (\pi; \calD)}

\newcommand{\dr}{\hat{V}_{\mathrm{DR}} (\pi; \calD, \hat{q})}
\newcommand{\dm}{\hat{V}_{\mathrm{DM}} (\pi; \calD, \hat{q})}

\newcommand{\mse}{\mathrm{MSE} (\hat{V}(\pi;\calD))}
\newcommand{\bias}{\mathrm{Bias} (\hat{V}(\pi;\calD))}
\newcommand{\biasoffcem}{\mathrm{Bias} ( \hat{V}_{\mathrm{OffCEM}} (\pi;\calD) )}
\newcommand{\var}{\mV_{\calD} \big[ \hat{V}(\pi;\calD) \big]}
\newcommand{\mseratioips}{\mathrm{MSE}(\hat{V}_{\mathrm{OffCEM}})/\mathrm{MSE}(\hat{V}_{\mathrm{IPS}})}
\newcommand{\mseratiodr}{\mathrm{MSE}(\hat{V}_{\mathrm{OffCEM}})/\mathrm{MSE}(\hat{V}_{\mathrm{DR}})}
\newcommand{\mseratiomips}{\mathrm{MSE}(\hat{V}_{\mathrm{OffCEM}})/\mathrm{MSE}(\hat{V}_{\mathrm{MIPS}})}
\icmltitlerunning{Off-Policy Evaluation for Large Action Spaces via Conjunct Effect Modeling}

\begin{document}

\twocolumn[
\icmltitle{Off-Policy Evaluation for Large Action Spaces\\ via Conjunct Effect Modeling} 

\begin{icmlauthorlist}
\icmlauthor{Yuta Saito}{cornell}
\icmlauthor{Qingyang Ren}{cornell}
\icmlauthor{Thorsten Joachims}{cornell}
\end{icmlauthorlist}
\icmlaffiliation{cornell}{Department of Computer Science, Cornell University, Ithaca, NY, USA}
\icmlcorrespondingauthor{Yuta Saito}{ys552@cornell.edu}
\icmlcorrespondingauthor{Thorsten Joachims}{tj@cs.cornell.edu}

\vskip 0.3in
]

\printAffiliationsAndNotice{} 

\begin{abstract}
We study off-policy evaluation (OPE) of contextual bandit policies for large discrete action spaces where conventional importance-weighting approaches suffer from excessive variance. To circumvent this variance issue, we propose a new estimator, called \textit{OffCEM}, that is based on the \textit{conjunct effect model} (CEM), a novel decomposition of the causal effect into a cluster effect and a residual effect. OffCEM applies importance weighting only to action clusters and addresses the residual causal effect through model-based reward estimation. We show that the proposed estimator is unbiased under a new condition, called \textit{local correctness}, which only requires that the residual-effect model preserves the relative expected reward differences of the actions within each cluster. To best leverage the CEM and local correctness, we also propose a new two-step procedure for performing model-based estimation that minimizes bias in the first step and variance in the second step. We find that the resulting OffCEM estimator substantially improves bias and variance compared to a range of conventional estimators. Experiments demonstrate that OffCEM provides substantial improvements in OPE especially in the presence of many actions.
\end{abstract}

\section{Introduction} \label{sec:intro}
Logged feedback is widely available in intelligent systems for a growing range of applications from media streaming to precision medicine. Many of these systems interact with their users through the \textit{contextual bandit} process, where a policy repeatedly observes a context, takes an action, and observes the reward for the chosen action. A common counterfactual question in this setting is that of \textit{off-policy evaluation} (OPE): how well would a new policy have performed, if it had been deployed instead of a logging policy? The ability to accurately answer this question using previously logged feedback data is of great practical importance, as it avoids costly and time-consuming online experiments~\cite{dudik2014doubly,su2020doubly}.

While we already have practical OPE estimators for small action spaces~\cite{dudik2014doubly,swaminathan2015batch,wang2017optimal,farajtabar2018more,su2019cab,su2020doubly,metelli2021subgaussian}, the effectiveness of these estimators degrades for large action spaces, which are common in many potential applications of OPE with thousands or millions of actions (e.g., movies, songs, products). In particular, when the number of actions is large, existing estimators -- most of which are based on importance weighting -- can collapse due to extreme variance caused by large importance weights~\cite{saito2022off}.

To alleviate the variance problem caused by large action spaces, this paper introduces the \textit{\textbf{C}onjunct \textbf{E}ffect \textbf{M}odel (\textbf{CEM})}, which decomposes the expected reward into a \textit{cluster effect} and a \textit{residual effect}. The cluster effect considers the realistic situation where some clustering of the actions already accounts for much of the reward variation (e.g., movie genres). The \textit{residual effect} only needs to model what causal effects remain from individual actions (e.g. movies) that are not already modeled by the cluster effect. This is a substantially more general formulation compared to the Marginalized Inverse Propensity Score (MIPS) estimator of~\citet{saito2022off}, which completely ignores the residual effect and assumes no direct effect for individual actions. Based on our new model of the reward function, we propose a novel estimator, called \textit{\textbf{Off}-policy evaluation estimator based on the \textbf{C}onjunct \textbf{E}ffect \textbf{M}odel (\textbf{OffCEM})}, which applies importance weighting over the action clusters to unbiasedly estimate the cluster effect while dealing with the residual effect via model-based estimation to avoid extreme variance. We first show that OffCEM is unbiased under a new condition called \textit{local correctness}, which only requires that the model-based residual-effect estimation preserves the relative value difference of the actions within each action cluster. To best leverage the structure of the CEM, we also provide a new learning procedure for a model-based estimator that directly optimizes the bias and variance of OffCEM. In particular, we propose to optimize regression models via a two-step procedure where the first-step minimizes the bias by optimizing local correctness and the second-step minimizes the variance by optimizing a baseline value function for each action cluster. 

Furthermore, we provide a thorough statistical comparison against a range of conventional estimators. In particular, we show that our estimator has a lower variance than Inverse Propensity Score (IPS), MIPS, and Doubly Robust (DR)~\cite{dudik2014doubly} because OffCEM considers importance weighting with respect to the action cluster space, which is much more compact than the original action space. Moreover, OffCEM often has a lower bias than MIPS because it explicitly addresses the residual effect via model-based estimation. Comprehensive experiments on synthetic and extreme classification data demonstrate that OffCEM can be substantially more effective than conventional estimators, improving the bias-variance trade-off particularly for challenging problems with many actions.

\section{Background} \label{sec:background}
We begin with a formal definition of OPE in the contextual bandit setting. 
We also describe and discuss the limitations of IPS, DR, and MIPS as the benchmark estimators.\footnote{Appendix~\ref{app:related} provides an extensive discussion of related work.}

\subsection{Off-Policy Evaluation} \label{sec:ope}
We formulate OPE following the general contextual bandit process, where a decision maker repeatedly observes a context $x \in \calX$ drawn i.i.d.\ from an unknown distribution $p(x)$. Given context $x$, a possibly stochastic \textit{policy} $\pi(a|x)$ chooses action $a$ from a finite action space denoted as $\calA$. The reward $r \in [0, r_{\mathrm{max}}]$ is then sampled from an unknown distribution $p(r|x,a)$, and we use $q(x,a) := \mE [r|x,a]$ to denote the expected reward given context $x$ and action $a$. We define the \textit{value} of $\pi$ as the key performance measure:
\begin{align*}
    \trueV := \mE_{p(x) \pi (a | x) p(r|x,a)} [r] = \mE_{p(x) \pi (a | x) } [q (x,a)]  
\end{align*}

The logged bandit data we use is of the form $\calD := \{(x_i,a_i,r_i)\}_{i=1}^n,$ which contains $n$ independent observations drawn from the logging policy $\pi_0$ as $(x,a,r) \sim p(x)\pi_0(a|x)p(r|x,a)$.  
In OPE, we aim to design an estimator $\hat{V}$ that can accurately estimate the value of a target policy $\pi$ (which is different from $\pi_0$) using only $\calD$. We measure the accuracy of $\hat{V}$ by its mean squared error (MSE)
\begin{align*}
    \mse 
    &:= \mE_{\calD} \Big[ \big(\trueV - \hat{V} (\pi; \calD) \big)^2 \Big],\\
    & = \bias^2 + \var,
\end{align*}
where $\mE_{\calD}[\cdot]$ denotes the expectation over the logged data. The bias and variance of $\hat{V}$ are defined respectively as
\begin{align*}
    \bias &:= \mE_{\calD}[\hat{V} (\pi; \calD)] - \trueV ,  \\
    \var & := \mE_{\calD} \Big[ \big(  \hat{V} (\pi; \calD) - \mE_{\calD} [\hat{V} (\pi; \calD)] \big)^2 \Big].
\end{align*}

\subsection{Limits of IPS and DR}
Our first benchmark is the IPS estimator, which forms the basis of many other OPE estimators ~\cite{dudik2014doubly,wang2017optimal,su2019cab,su2020doubly,metelli2021subgaussian}. IPS estimates the value of $\pi$ by re-weighting the observed rewards as
\begin{align*}
    \ips := \meanN \frac{\pi(a_i\,|\,x_i)}{\pi_0 (a_i\,|\,x_i)} r_i = \meanN w(x_i, a_i) r_i,
\end{align*}
where $w(x,a) := \pi(a\,|\,x) / \pi_0(a\,|\,x)$ is the \textit{(vanilla) importance weight}.

This estimator is unbiased (i.e., $ \mE_{\calD} [\ips] = \trueV $) under the following common support assumption. 
\begin{assumption} (Common Support) \label{assumption:common_support}
The logging policy $\pi_0$ is said to have common support for policy $\pi$ if $\pi (a \,|\, x) > 0 \rightarrow \pi_0(a \,|\, x) > 0$ for all $a \in \calA$ and $x \in \calX$.
\end{assumption}

The unbiasedness of IPS is indeed desirable. However, a critical issue is that its variance can be excessive, particularly when there is a large number of actions. The DR estimator~\citep{dudik2014doubly} has been a popular technique to reduce the variance by incorporating a model-based reward estimator $\hat{q}(x,a) \approx q(x,a)$ as follows. 
\begin{align*}
    &\dr \\
    &:= \meanN \left\{ w(x_i, a_i ) (r_i - \hat{q}(x_i,a_i)) + \hat{q}(x_i,\pi) \right\},
\end{align*}
where $\hat{q}(x,\pi):= \mE_{\pi(a|x)} [\hat{q}(x,a)]$. 
DR often improves the MSE of IPS by reducing the variance. However, its variance can still be extremely large due to vanilla importance weighting, which leads to substantial accuracy deterioration in large action spaces~\citep{saito2022off}. This issue of IPS and DR can be seen by calculating their variance
\begin{align}
    n \mV_{\calD} \big[  \dr \big] 
    & =  \mE_{p(x)\pi_0(a|x)} [ w(x,a)^2 \sigma^2 (x,a) ] \notag \\
    & + \mE_{p(x)} \left[  \mV_{\pi_0(a|x)} [ w(x,a) \Delta_{q,\hat{q}} (x,a) ] \right] \notag\\ 
    & + \mV_{p(x)} \left[  \mE_{\pi(a|x)} [ q (x,a) ] \right], \label{eq:dr_variance}
\end{align}
where $ \sigma^2 (x,a) := \mV [r|x,a] $ and $\Delta_{q,\hat{q}} (x,a) := q(x,a)-\hat{q}(x,a)$. Note that the variance of IPS can be obtained by setting $\hat{q}(x,a) = 0$. The variance reduction of DR comes from the second term where $\Delta_{q,\hat{q}} (x,a)$ is smaller than $q(x,a)$ if $\hat{q}(x,a)$ is reasonably accurate. However, we can also see that the variance contributed by the first term can be extremely large for both IPS and DR when the reward is noisy and the weights $w(x,a)$ have a wide range, which occurs when $\pi$ assigns large probabilities to actions that have low probability under $\pi_0$. This is often the case when there are many actions and the logging policy $\pi_0$ aims to guarantee \textit{universal support} (i.e., $\pi_0(a\,|\,x)>0$ for all $a$ and $x$).

\subsection{Limits of The MIPS Estimator} \label{sec:aaa}
To address the excessive variance of IPS and DR, \citet{saito2022off} assume the existence of \textit{action embeddings} in the logged data. There are many cases where we have access to such prior information about the actions. For example, movies are characterized by genres (e.g., adventure, documentary), director, or actors. The idea of MIPS~\citep{saito2022off} is to utilize this auxiliary information about the actions to perform OPE more efficiently.

More formally, suppose we are given a $d_e$-dimensional \textit{action embedding} $e \in \calE \subseteq\mathbb{R}^{d_e}$ for each action $a$, where the embedding is drawn i.i.d.\ from some unknown distribution $p(e\,|\,x,a)$. If the embedding is defined by predefined category information, for instance, then it is independent of the context and is deterministic given the action. However, if the action embedding is, for example, a product price generated by some personalized and randomized pricing algorithm, it becomes continuous, stochastic, and context-dependent.

Leveraging the predefined action embeddings, \citet{saito2022off} propose the following MIPS estimator.
\begin{align*}
    \mips := \meanN \!\frac{\pi(e_i\,|\,x_i)}{\pi_0(e_i\,|\,x_i)} r_i = \meanN w(x_i, e_i) r_i,
\end{align*}
where the logged dataset $\calD= \{(x_i, a_i, e_{i}, r_i)\}_{i=1}^n$ now contains action embeddings for each data and
\begin{align*}
    w(x,e) := \frac{\pi(e\,|\,x)}{\pi_0(e\,|\,x)} = \frac{\sumA p(e\,|\,x, a) \pi(a\,|\,x)}{\sumA p(e\,|\, x, a) \pi_0(a\,|\,x)}
\end{align*}
is the \textit{marginal importance weight} based on marginal embedding distribution $\pi(e\,|\,x) := \sumA p(e\,|\, x, a) \pi(a\,|\,x) $.

This estimator is unbiased (i.e., $ \mE_{\calD} [\mips] = \trueV $) under the following assumptions~\citep{saito2022off}.
\begin{assumption} (Common Embedding Support) \label{assumption:common_embed_support}
The logging policy $\pi_0$ is said to have common embedding support for policy $\pi$ if $\pi (e\,|\,x) > 0 \rightarrow \pi_0(e\,|\,x) > 0$ for all $e$ and $x$.
\end{assumption}
\vspace{1mm}
\begin{assumption} (No Direct Effect) \label{assumption:no_direct_effect}
Action $a$ has no direct effect on the reward $r$ given the embedding $e$ if $a \perp r \mid x,e$. 
\end{assumption}

Assumption~\ref{assumption:common_embed_support} is a weaker version of Assumption~\ref{assumption:common_support}, requiring common support only with respect to the action embedding space. Assumption~\ref{assumption:no_direct_effect} is about the quality of the given action embeddings and requires that every possible effect of $a$ on $r$ be fully mediated by embedding $e$. 

Even though MIPS often improves the MSE over existing estimators such as IPS and DR by utilizing action embeddings, we argue that MIPS can still suffer from substantial bias or variance in many realistic situations. First, if the given action embeddings are high-dimensional and fine-grained, MIPS becomes very similar to IPS, producing extreme variance. This is because, in such cases, the embedding distribution $p(e\,|\,x,a)$ is near-deterministic and thus the marginal importance weight becomes close to the vanilla importance weight, i.e., $w(x,a) \approx w(x,e)$. We can reduce the variance of MIPS by performing action feature selection as described in~\citet{saito2022off}, but this may produce a large amount of bias by violating Assumption~\ref{assumption:no_direct_effect}. This bias-variance dilemma of MIPS motivates the development of a new framework and estimator for large action spaces.

\begin{figure}[t]
\centering
\includegraphics[clip, width=8cm]{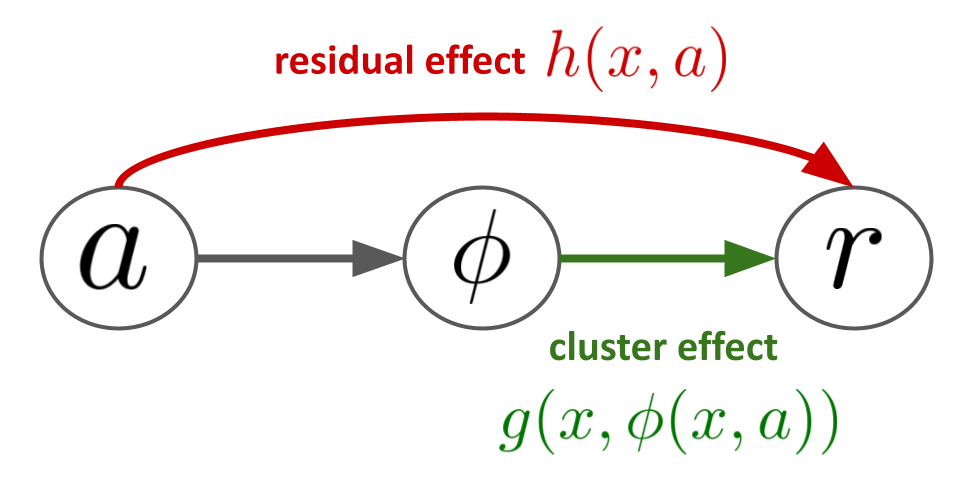}
\caption{The Conjunct Effect Model (CEM)} \vspace{-5mm} \label{fig:cem}
\raggedright
\end{figure}

\section{The Conjunct Effect Model and OffCEM Estimator} \label{sec:method}
The following proposes a new estimator that circumvents the challenges of MIPS. The key idea is to decompose the expected reward into the \textit{cluster effect} and \textit{residual effect} rather than making the no direct effect assumption, which might be unrealistic and thus cause the dilemma. Specifically, given some action clustering function $\phi: \calX \times \calA \rightarrow \calC$, which may be learned from log data, where $\calC$ is an action cluster space (typically $|\calC| \ll |\calA|$), we consider the following \textit{conjunct effect model} (CEM), which decomposes the expected reward function into two separate effects.

\begin{tcolorbox}
\textbf{The Conjunct Effect Model (CEM):}
\begin{align}
    q(x,a) = \underbrace{g(x,\phi(x,a))}_{\textit{cluster effect}} + \underbrace{h(x,a)}_{\textit{residual effect}} \label{eq:cem}
\end{align}
\end{tcolorbox}

For example, in a movie recommendation problem, the cluster effect could capture the relevance of each genre to the users, and the residual effect models how each movie is better or worse than the overall genre preference. In the simplest case, a non-personalized residual effect can model that some movies in each genre are generally better than others, or it can model a personalized effect that a specific user likes a particular actor. Note that Assumption~\ref{assumption:no_direct_effect} requires no residual effect for MIPS, i.e., $h(x,a)=0$ for all $(x,a)$, so our CEM formulation is strictly more general than that of \citet{saito2022off}.\footnote{Note that we consider action clusters as a particular type (one-dimensional and discrete) of low-dimensional action embeddings for brevity of exposition and analysis. By doing so, the CEM can be formulated using the standard logged dataset of the form: $\calD = \{(x_i,a_i,r_i)\}_{i=1}^n$, which does not include action embeddings $e$. However, our framework can be extended to a more general formulation, as shown below: \begin{align} q(x,a,e) = \underbrace{g(x,\phi(x,e))}_{\textit{embedding effect}} + \underbrace{h(x,a,e)}_{\textit{residual effect}} \label{eq:general_cem}, \end{align} where $e$ is a raw action embedding recorded in the dataset and $\phi: \calX \times \calE \rightarrow \mathbb{R}^{d_{\phi}}$ is a $d_{\phi}$-dimensional (lower-dimensional) latent representation of the action.}

Our proposed estimator, which we call the \textbf{OffCEM} estimator, overcomes the limitations of MIPS by appropriately selecting the estimation strategies for the cluster and residual effect based on the CEM as follows.
\begin{align}
    &\offcem  \label{eq:OffCEM} \\ 
    &\!\! := \meanN \bigg\{ w(x_i,\phi(x_i,a_i)) (r_i - \hat{f}(x_i,a_i)) + \hat{f}(x_i,\pi) \bigg\}, \notag
\end{align}
where $\hat{f}(x,\pi) := \mE_{\pi(a|x)}[\hat{f}(x,a)]$ and we define the \textit{cluster importance weight} as
\begin{align*}
    w(x,c) := \frac{\pi(c\,|\,x)}{\pi_0(c\,|\,x)} = \frac{\sumA \mathbb{I} \{ \phi(x,a) = c \} \pi(a\,|\,x) }{\sumA \mathbb{I} \{ \phi(x,a) = c \} \pi_0(a\,|\,x) },
\end{align*}
for $c\in \calC$. The first term of OffCEM estimates the cluster effect via cluster importance weighting and the second term deals with the residual effect via the regression model $\hat{f}$.\footnote{Note that our estimator may look similar to DR at first glance, but ours is derived from applying different estimation strategies between the two terms in the CEM and this design principal is substantially different from that of DR.}

Intuitively, our estimator is expected to perform better than a range of typical estimators for large action spaces. First, OffCEM performs importance weighting with respect to the action cluster space and thus is expected to have a much lower variance than IPS, DR, and MIPS, which apply importance weighting with respect to the original action space (IPS and DR) or potentially high-dimensional action embeddings (MIPS). Moreover, OffCEM can have a lower bias than MIPS by dealing with the residual effect via model-based estimation (the second term) rather than ignoring it as in MIPS. Note that a seemingly natural choice of the regression model $\hat{f}$ is a direct estimate of the residual effect, i.e., $\hat{f} \approx h(x,a)$, but we later show that there is a more refined \textit{two-step} procedure to optimize the regression model to best leverage the structure of our conjunct effect model. 

The following analyzes the statistical properties of OffCEM, showing that it is unbiased under a new condition called local correctness and that it has a much more favorable bias-variance tradeoff compared to the conventional estimators. We will also discuss how to optimize the regression model $\hat{f}$ based on our statistical analysis.

\subsection{Theoretical Analysis}
First, we formally introduce the \textit{local correctness} assumption and show that OffCEM is unbiased under it. 

\begin{assumption} (Local Correctness) \label{assumption:local_correctness}
A regression model $\hat{f}(\cdot,\cdot)$ and action clustering function $\phi(\cdot,\cdot)$ satisfy local correctness if the following holds true:
\begin{align}
    & \Delta_q(x,a,b) = \Delta_{\hat{f}} (x,a,b) \label{eq:local_correctness},
\end{align}
for all $x\in\calX$ and $a,b\in\calA$ with $\phi(x,a)=\phi(x,b)$, where $\Delta_q(x,a,b) := q(x,a) - q(x,b)$ is the difference in the expected rewards between the actions $a$ and $b$ given context $x$, which we call the \textit{relative value difference} of the actions. $\Delta_{\hat{f}} (x,a,b):= \hat{f}(x,a) - \hat{f}(x,b)$ is an estimate of the relative value difference between $a$ and $b$ based on $\hat{f}$.
\end{assumption}

Assumption~\ref{assumption:local_correctness} only requires that the regression model $\hat{f}$ should correctly preserve the \textit{relative value difference} $\Delta_q(x,a,b)$ between the actions within each action cluster $c$. Thus, to satisfy local correctness, we need not be able to accurately estimate the absolute value of the reward function, which is typically required for model-based estimators and control variates. Note also that local correctness does not require any particular relationship across different clusters. For instance, suppose there are three movies $a=1,2,3$ where $a=1$ and $a=2$ belong to the first cluster $c=1$ while $a=3$ belongs to the second cluster $c=2$. Then, local correctness only requires a regression model $\hat{f}$ to preserve the relative value difference between $a=1$ and $a=2$, but $\hat{f}$ does not need to learn any particular relationship between $a=1,2$ (the first cluster) and $a=3$ (the second cluster). Appendix~\ref{app:local_correctness} provides more specific examples of locally correct regression models.

The following shows that local correctness is indeed a new requirement for an unbiased OPE based on our framework.
\begin{proposition} \label{prop:unbiased}
Under Assumptions~\ref{assumption:common_embed_support} (wrt $\calE = \calC$) and~\ref{assumption:local_correctness}, OffCEM is unbiased, i.e., $\mE_{\mathcal{D}}[\offcem] = \trueV$. See Appendix~\ref{app:unbias} for the proof.
\end{proposition}

Proposition~\ref{prop:unbiased} shows that OffCEM is unbiased even when MIPS is biased due to violations of Assumption~\ref{assumption:no_direct_effect}, if we can merely learn the relative value difference within each cluster $c$. Beyond unbiasedness, the following also characterizes the magnitude of the bias of OffCEM when Assumption~\ref{assumption:local_correctness} is violated.

\begin{theorem} (Bias of OffCEM) \label{thm:bias}
If Assumption~\ref{assumption:common_embed_support} (wrt $\calE = \calC$) is true, but Assumption~\ref{assumption:local_correctness} is violated, OffCEM has the following bias for some given regression model $\hat{f}(\cdot,\cdot)$ and clustering function $\phi(\cdot,\cdot)$. \vspace{-3mm}
\begin{align*}
    &\biasoffcem \\
    &= \mE_{p(x)\pi(c|x)} 
    \Bigg[ 
        \sum_{a < b:\phi(x,a)=\phi(x,b)=c} \!\!\!  \pi_0(a\,|\,x,c) \pi_0(b\,|\,x,c) \\
        & \qquad \qquad \qquad \qquad \quad \times (\Delta_q(x,a,b) - \Delta_{\hat{f}}(x,a,b)) \\ 
        & \qquad \qquad \qquad \qquad \quad \times \left(\frac{\pi(b\,|\,x,c)}{\pi_0(b\,|\,x,c)} - \frac{\pi(a\,|\,x,c)}{\pi_0(a\,|\,x,c)}\right)
    \Bigg],
\end{align*}
where $a,b \in \calA$ and $\pi_0(a\,|\,x,c) = \pi_0(a\,|\,x)\mathbb{I}\{\phi(x,a)=c\}/\pi_0(c\,|\,x) $. See Appendix~\ref{app:bias} for the proof.
\end{theorem}

Theorem~\ref{thm:bias} shows that three factors contribute to the bias of OffCEM when Assumption~\ref{assumption:local_correctness} is violated. The first factor is the \textit{entropy of the logging policy within each action cluster}. 
When action $a$ is near deterministic given context $x$ and cluster $c$, $\pi_0(a\,|\,x,c)$ becomes close to zero or one, making $\pi_0(a\,|\,x,c)\pi_0(b\,|\,x,c)$ close to zero. This suggests that even if Assumption~\ref{assumption:local_correctness} is violated, a sufficiently large cluster space still enables nearly unbiased estimation with OffCEM. The second factor is the \textit{accuracy of the regression model $\hat{f}$ with respect to the relative value difference}, which is quantified by $\Delta_q(x,a,b) - \Delta_{\hat{f}}(x,a,b)$. When $\hat{f}$ preserves the relative value difference of the actions within each cluster well, the second factor becomes small and so does the bias of OffCEM. This also suggests that, in an ideal case when Assumption~\ref{assumption:local_correctness} is satisfied, OffCEM is unbiased for any target policy, recovering Proposition~\ref{prop:unbiased}. The final factor is the \textit{relative similarity between logging and target policy within each action cluster} as quantified by $\frac{\pi(b\,|\,x,c)}{\pi_0(b\,|\,x,c)} - \frac{\pi(a\,|\,x,c)}{\pi_0(a\,|\,x,c)}$, which suggests that the bias of OffCEM becomes small if the target and logging policies are similar within each action cluster. Note that this still allows the target and logging policy to be different in how they choose between clusters for the third factor to diminish and thus drive the overall bias to zero.

Next, we analyze the variance of OffCEM, which provides additional insights as to how we should optimize the regression model $\hat{f}$ to best exploit the CEM from Eq.~\eqref{eq:cem}. 
\begin{proposition} (Variance of OffCEM) \label{prop:variance}
Under Assumptions~\ref{assumption:common_embed_support} (wrt $\calE = \calC$) and~\ref{assumption:local_correctness}, OffCEM has the following variance.
\begin{align}
    & n \mV_{\calD} \big[  \offcem \big]  \notag \\
    & =  \mE_{p(x)\pi_0(a|x)} [ w(x,\phi(x,a))^2 \sigma^2 (x,a) ] \notag \\
    & \quad + \mE_{p(x)} \left[  \mV_{\pi_0(a|x)} [ w(x,\phi(x,a)) \Delta_{q,\hat{f}}(x,a) ] \right] \notag \\
    & \quad + \mV_{p(x)} \left[  \mE_{\pi(a|x)} [q(x,a)] \right], \label{eq:OffCEM_variance}
\end{align}
where $\Delta_{q,\hat{f}}(x,a):= q(x,a) - \hat{f}(x,a)$ is the estimation error of $\hat{f}(x,a)$ against the expected reward function $q(x,a)$. See Appendix~\ref{app:variance} for the proof.
\end{proposition}
Proposition~\ref{prop:variance} shows that the variance of OffCEM depends only on the cluster importance weight rather than the vanilla importance weight, implying reduced variance compared to IPS and DR (c.f., Eq.~\eqref{eq:dr_variance}). Moreover, when action embeddings $e$ are high-dimensional and fine-grained, the marginal importance weight of MIPS becomes very similar to the vanilla importance weight, and thus the variance of OffCEM is expected to be smaller than that of MIPS in such a situation. Moreover, Eq.~\eqref{eq:OffCEM_variance} implies that we can improve the variance of OffCEM by minimizing $|\Delta_{q,\hat{f}}(x,a)|$, which suggests an interesting \textit{two-step} strategy for optimizing the regression model $\hat{f}$ as described in the next subsection.

\subsection{A Two-Step Regression Procedure} \label{sec:two-step_regression}
The analysis in the previous section suggests how we should optimize the regression model $\hat{f}$ for a given clustering function $\phi$.\footnote{This clustering can be produced by a standard clustering algorithm, but the development of a refined clustering procedure that more explicitly optimizes the statistical properties of OffCEM is an interesting direction for future work.} More specifically, Theorem~\ref{thm:bias} showed that the accuracy of estimating the relative value difference $\Delta_q(x,a,b)$ (i.e., satisfaction of local correctness) characterizes the bias of OffCEM while we should minimize $|\Delta_{q,\hat{f}}(x,a)|$ for minimizing its variance as implied in Proposition~\ref{prop:variance}. Therefore, we propose the following two-step procedure for optimizing a regression model where the first-step corresponds to bias minimization while the second-step aims to minimize the variance of the resulting estimator.
\begin{enumerate}
    \item \textbf{Bias Minimization Step}: Optimize the pairwise regression function $\hat{h}_{\theta}: \calX \times \calA \rightarrow \mathbb{R}$, parameterized by $\theta$, to approximate $\Delta_q(x,a,b)$ via
        \begin{align*}
            \min_{\theta} \!\! \sum_{(x,a,b,r_a,r_b) \in \calD_{pair}} \!\! \ell_h \left(r_{a} - r_{b}, \hat{h}_{\theta} (x,a) -  \hat{h}_{\theta} (x,b) \right).
        \end{align*}
    \item \textbf{Variance Minimization Step}: Optimize the baseline value function $\hat{g}_{\psi}: \calX \times \calC \rightarrow \mathbb{R}$, parameterized by $\psi$, to minimize $\Delta_{q,\hat{f}}(x,a)$ given $\hat{f} = \hat{g}_{\psi} + \hat{h}_{\theta}$ via
        \begin{align*}
            \min_{\psi} \sum_{(x,a,r) \in \calD} \ell_g \left(r - \hat{h}_{\theta} (x,a), \hat{g}_{\psi} (x,\phi(x,a)) \right).
        \end{align*}
\end{enumerate}
 $\ell_h, \ell_g: \mathbb{R} \times \mathbb{R} \rightarrow \mathbb{R}_{\ge 0}$ are some appropriate loss functions such as cross-entropy or squared loss. As suggested in our analysis, $\hat{h}_{\theta}(x,a)$ fully characterizes the bias of OffCEM, and thus the second step can focus entirely on variance minimization by optimizing the baseline value function $\hat{g}_{\psi}(x,\phi(x,a))$, which does not affect the bias. After performing the two-step regression, we construct our regression model as $\hat{f}_{\theta,\psi}(x,a) = \hat{g}_{\psi}(x,\phi(x,a)) + \hat{h}_{\theta}(x,a)$.

Note that $\calD_{pair}$ is a dataset augmented for performing the bias minimization step, which is defined as
\begin{align*}
    &\calD_{pair} \\
    &:= \! \Bigg\{ (x,a,b,r_a,r_b) \! \mid
        \begin{array}{ll}
            \quad (x_a,a,r_a), (x_b,b,r_b) \in \calD \! \\ 
            \! x=x_a=x_b, \phi(x,a) = \phi(x,b) \!
        \end{array}
    \Bigg\},
\end{align*}
where we assume the context space is finite here. If it is difficult to find a pair of observed actions that have the same cluster, we can instead define $\calD_{pair}$ as \begin{align*}
        \calD_{pair} 
        &:=  \! \bigg\{ (x,a,b,r_a,r_b)  \! \mid
            \begin{array}{ll}
                \! (x_a,a,r_a), (x_b,b,r_b) \in \calD \\ \quad \quad x=x_a=x_b
            \end{array}
        \bigg\},
    \end{align*} 
which may perform better empirically due to increased sample size. Note that, even if this two-step procedure is infeasible due to insufficient pairwise data, we can still perform a conventional regression for the expected absolute reward to learn the parameterized function $\hat{f}_{\omega}: \calX \times \calA \rightarrow \mathbb{R}$ via:
\begin{align}
    \min_{\omega} \sum_{(x,a,r) \in \calD } \ell_f \left(r, \hat{f}_{\omega} (x,a) \right), \label{eq:one_step_reg}
\end{align}
and then use $\hat{f}_{\omega}$ in Eq.~\eqref{eq:OffCEM}. Even for such a conventionally trained regression model $\hat{f}_{\omega}$, the OffCEM estimator still has advantages over the benchmark estimators. In particular, cluster importance weighting still provides improved variance and $\hat{f}_{\omega}$ is still preferable over ignoring the residual effect. In the following experiments, we empirically evaluate the effectiveness of our two-step regression procedure, but we also find that OffCEM performs better than existing estimators even for conventionally trained $\hat{f}_{\omega}$ via Eq.~\eqref{eq:one_step_reg}.

\begin{figure*}[th]
\centering
\begin{minipage}{\hsize}
    \begin{center}
        \includegraphics[clip, width=17cm]{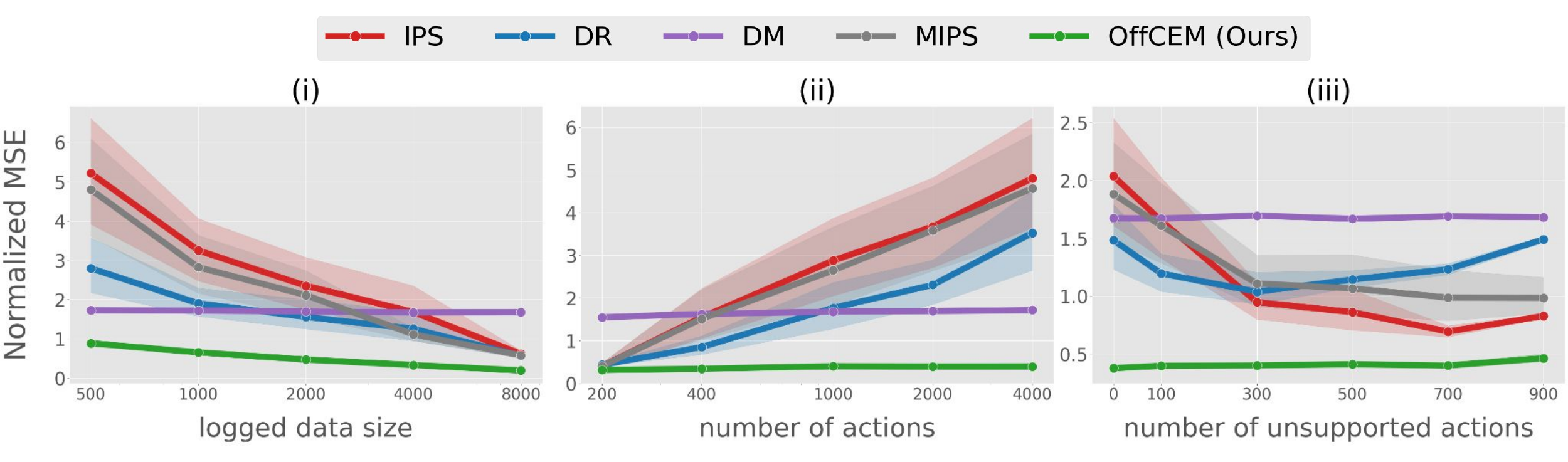}
    \end{center}
    \vspace{-3mm}
    \caption{Comparison of the estimators' MSE (normalized by the true value $V(\pi)$) with (i) \textbf{varying logged data sizes ($n$)}, (ii) \textbf{varying numbers of actions ($|\calA|$)}, and (iii) \textbf{varying numbers of unsupported actions ($|\calU|$)} in the synthetic experiment.} \vspace{-2mm}
    \label{fig:main}
\end{minipage}
\end{figure*}

\begin{table*}[th]
\caption{Improvements in MSE provided by OffCEM against the baselines in the synthetic experiment. A smaller value indicates a larger improvement by OffCEM. (default experiment parameters: $n=3,000, |\calA|=1,000$, and $|\calU|=0$)} \label{tab:improvement}
\vspace{2mm}
\centering
\scalebox{1.05}{
\begin{tabular}{c|cc|cc|cc}
\toprule
 & $n = 500$ & $n = 8,000$ & $|\calA|=200$ & $|\calA|=4,000$ & $|\calU|=0$ & $|\calU|=900$ \\\midrule 
$ \mseratioips $ & 0.169 & 0.314 & 0.767 & 0.082 & 0.185 & 0.560 \\
$ \mseratiomips $ & 0.184 & 0.335 & 0.770 & 0.086 & 0.200 & 0.471 \\
$ \mseratiodr $ & 0.317 & 0.339 & 0.720 & 0.112 & 0.254 & 0.311 \\
\bottomrule
\end{tabular}
}
\end{table*}

\begin{figure*}[th]
\centering
\begin{minipage}{\hsize}
    \vspace{2mm}
    \begin{center}
        \includegraphics[clip, width=17cm]{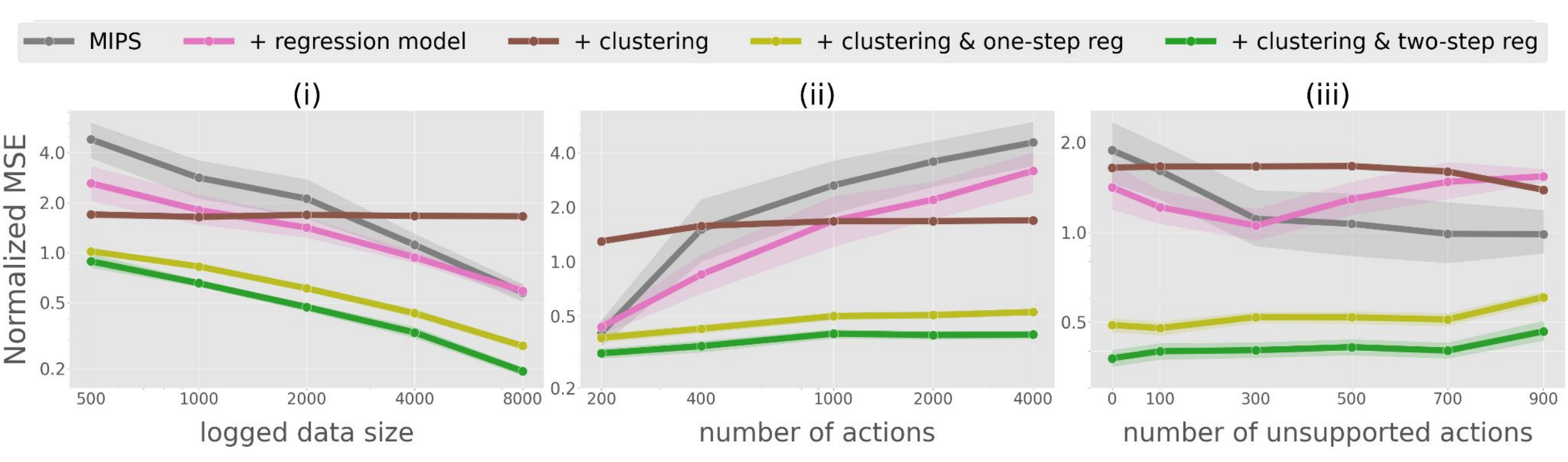}
    \end{center}
    \vspace{-2mm}
    \caption{Ablation analysis to evaluate the contribution of the key components of OffCEM with (i) \textbf{varying logged data sizes ($n$)}, (ii) \textbf{varying numbers of actions ($|\calA|$)}, and (iii) \textbf{varying numbers of unsupported actions ($|\calU|$)}.}
    \label{fig:ablation}
\end{minipage}
\end{figure*}
\section{Empirical Evaluation} \label{sec:empirical}
We first evaluate OffCEM on synthetic data to identify the situations where it enables a more accurate OPE. Second, we validate real-world applicability of OffCEM on extreme classification datasets, which can be converted into bandit problems with large action spaces. Our experiments are conducted using the \textit{OpenBanditPipeline} (OBP)\footnote{\href{https://github.com/st-tech/zr-obp}{\blue{https://github.com/st-tech/zr-obp}}}, an open-source software for OPE provided by~\citet{saito2021open}. The experiment code is publicly available on GitHub: \href{https://github.com/usaito/icml2023-offcem}{\blue{https://github.com/usaito/icml2023-offcem}}.

\subsection{Synthetic Data} \label{sec:synthetic}
For the first set of experiments, we create synthetic datasets to be able to compare the estimates to the ground-truth value of the target policies. Our setup imitates a recommender system, where each user repeatedly interacts with the system to provide data useful for the two-step regression method from Section~\ref{sec:two-step_regression}. Specifically, we first define 200 unique users whose 10-dimensional context vectors $x$ are sampled from the standard normal distribution. We then assign 10-dimensional categorical action embedding $e_a \in \calE$ to each action $a$ where each dimension of $\calE$ has a cardinality of 5.\footnote{This is higher-dimensional and finer-grained than the embeddings used in the synthetic experiment of \citet{saito2022off}.} We also cluster the actions based on their embeddings where the cluster assigned to action $a$ is denoted as $c_a (= \phi(a))$. We then synthesize the expected reward function as
\begin{align}
    q(x,a) = g(x,c_a ) + h_{c_a}(x,a) \label{eq:synthetic_reward}
\end{align}
Appendix~\ref{app:synthetic} defines $g(\cdot,\cdot)$ and $h_{\cdot}(\cdot,\cdot)$ in detail. We then sample the reward $r$ from a normal distribution with mean $q(x,a)$ and standard deviation $\sigma=3$.

We synthesize the logging policy $\pi_0$ by applying the softmax function to the expected reward function $q(x,a)$ as
\begin{align}
    \pi_0(a \,|\, x) =  \frac{\exp( \beta \cdot q(x,a))}{ \sum_{a' \in \calA} \exp( \beta \cdot q(x,a')) } , \label{eq:synthetic_logging}
\end{align}
where $\beta$ is a parameter that controls the optimality and entropy of the logging policy, and we use $\beta=-0.1$.

In contrast, the target policy $\pi$ is defined as
\begin{align}
    \pi(a \,|\, x) = (1 - \epsilon) \cdot \mathbb{I} \big\{a = \argmax_{a' \in \calA} q(x,a') \big\} + \frac{\epsilon}{|\calA|}, \label{eq:synthetic_target}
\end{align}
where the noise $\epsilon \in [0,1]$ controls the quality of $\pi$, and we set $\epsilon=0.2$ in the main text. Appendix~\ref{app:synthetic} provides additional results for varying values of $\epsilon$.

To summarize, we first sample a user and define the expected reward $q(x,a)$ as in Eq.~\eqref{eq:synthetic_reward}. We then sample discrete action $a$ from $\pi_0$ based on Eq.~\eqref{eq:synthetic_logging} where action $a$ is associated with a cluster $c_a$ and embedding vector $e_a$. The reward is then sampled from a normal distribution with mean $q(x,a)$. Iterating this procedure $n$ times generates logged data $\calD$ with $n$ independent copies of $(x,a,e_a,c_a,r)$ where the same user may appear multiple times in the logged data and thus the two-step procedure from Section~\ref{sec:two-step_regression} is applicable.

\subsubsection{Baselines}
We compare our estimator with the Direct Method (DM), IPS, DR, and MIPS. We optimize the regression model for OffCEM following the two-step procedure described in Section~\ref{sec:two-step_regression}. We use a neural network with 3 hidden layers along with 3-fold cross-fitting~\cite{newey2018crossfitting} to obtain $\hat{q}(x,a)$ for DR and DM, and $(\hat{h}_{\theta}, \hat{g}_{\psi})$ for OffCEM.

\subsubsection{Results}
Figures~\ref{fig:main} and~\ref{fig:ablation} show the MSE (normalized by $V(\pi)$) of the estimators computed over 300 simulations with different random seeds. Table~\ref{tab:improvement} shows the MSE of OffCEM relative to those of the baselines where a smaller value indicates a larger improvement by OffCEM. Note that we use $n=3,000$, $|\calA|=1,000$, and $|\calC|=50$ as default parameters.

\paragraph{Performance comparisons.}
First, Figure~\ref{fig:main} (i) compares the estimators’ performance when we vary the logged data size from 500 to 8,000. We observe that all estimators except for DM improve the MSE with increasing logged data sizes as expected, but OffCEM provides substantial improvements in MSE over IPS, DR, and MIPS, particularly when the logged data size is small. More specifically, Table~\ref{tab:improvement} shows that OffCEM provides approximately 83.0\% improvement (reduction) in MSE compared to IPS, 81.5\% improvement compared to MIPS, and 68.2\% improvement compared to DR when $n=500$. The improvements of OffCEM when $n=8,000$ are still substantial but slightly smaller. Note that MIPS performs almost identically to IPS since its marginal importance weight becomes similar to the vanilla importance weight given fine-grained embeddings. OffCEM also performs much better than DM, which suffers from high bias. These strong results of OffCEM are due to its almost zero bias and a variance similar to that of DM.\footnote{Appendix~\ref{app:synthetic} reports and discusses the bias-variance decomposition of the synthetic results.}

Next, Figure~\ref{fig:main} (ii) reports the estimators’ performance when we increase the number of actions from 200 to 4,000. We can see that OffCEM is appealing, particularly for large action spaces, where it outperforms IPS, DR, and MIPS by a larger margin than in small action spaces. Specifically, Table~\ref{tab:improvement} suggests that OffCEM improves IPS and MIPS by approximately 99.1\% and DR by 88.7\% when $|\calA| = 4,000$, which are substantially larger improvements compared to when $|\calA| = 200$. We can also see that MIPS suffers from large action spaces similarly to IPS since its marginal importance weight $w(x,e)$ becomes almost identical to the vanilla importance weight $w(x,a)$ of IPS in our synthetic environment. DR is slightly better than IPS and MIPS, but it still suffers from growing variance due to vanilla importance weighting when the number of actions becomes larger. These results suggest that cluster importance weighting is increasingly effective in reducing the variance in larger action spaces, while marginal importance weighting of MIPS fails to improve IPS given fine-grained action embeddings.

Finally, Figure~\ref{fig:main} (iii) evaluates the estimators’ performance when we increase the number of unsupported actions from 0 to 900, which introduces some bias in IPS, DR, and MIPS by violating Assumption~\ref{assumption:common_support}~\citep{sachdeva2020off}. We can see from the figure that OffCEM is robust to the violation of Assumption~\ref{assumption:common_support}, which is likely to occur in large action spaces, and is consistently more favorable compared to the baseline estimators in various numbers of unsupported actions $|\calU|$ (where $\calU := \{a \in \calA \,|\, \pi_0 (a \,|\, \cdot) = 0\}$ is the set of unsupported actions). This is due to the fact that OffCEM avoids producing large bias in the cluster effect estimate as long as some actions in each cluster are still supported, while IPS, DR, and MIPS drastically increase their bias. Note that a larger number of unsupported actions introduce a larger bias in IPS, DR, and MIPS, but their MSEs do not necessarily worsen due to reduced variance with a smaller number of supported actions where logging and target policies become relatively similar.

\paragraph{Ablation analysis.}
In addition to the performance comparisons, we perform an ablation study to investigate how effective each component of OffCEM is. 
For this purpose, in Figure~\ref{fig:ablation}, we gradually add the key components (cluster importance weighting and regression model) of OffCEM to MIPS and compare their MSE in various settings.\footnote{The precise definitions of each estimator compared in Figure~\ref{fig:ablation} can be found in Appendix~\ref{app:baselines}. In particular, \textbf{``+clustering"} uses only cluster importance weighting as $\meanN w(x_i,\phi(x_i,a_i)) r_i$ while \textbf{``+regression model"} adds only a one-step regression model $\hat{q}(x,a)$ to MIPS as $\meanN \left\{ w(x_i,e_i) (r_i - \hat{q}(x_i,a_i)) + \hat{q}(x_i,\pi) \right\}$. \textbf{``+clustering \& two-step reg"} and \textbf{``+clustering \& one-step reg"} are versions of OffCEM w/ or w/o two-step regression.}

We can see from Figure~\ref{fig:ablation} that combining cluster importance weighting and regression model is crucial for improving OPE in large action spaces. More specifically, using only cluster importance weighting (\textbf{``+clustering"}) ignores the residual effect and often produces large bias while using only a regression model (\textbf{``+regression model"}) produces high variance due to marginal importance weights. The result also suggests that combining only cluster importance weight and one-step regression as in Eq.~\eqref{eq:one_step_reg} is much better than  
\textbf{``+clustering"} and \textbf{``+regression model"}, and effective enough to greatly improve MIPS. But, we should perform two-step regression when feasible to provide further improvements, which we can see by comparing \textbf{``+clustering \& one-step reg"} and \textbf{``+clustering \& two-step reg"}.

\begin{table}[t]
\caption{Dataset Statistics} \label{tab:data_stats}
\vspace{2mm}
\centering
\scalebox{1.05}{
\begin{tabular}{c|ccc}
\toprule
\textbf{} & $n_{train}$ & $n_{test}$ & $|\calA|$ \\\midrule 
EUR-Lex 4K & 15,449 & 3,865 & 3,956 \\
Wiki10-31K & 14,146 & 6,616 & 30,938 \\
\bottomrule
\end{tabular}
}
\end{table}
\begin{figure*}[th]
\vspace{3mm}
\centering
\begin{minipage}{\hsize}
    \begin{center}
        \includegraphics[clip, width=15cm]{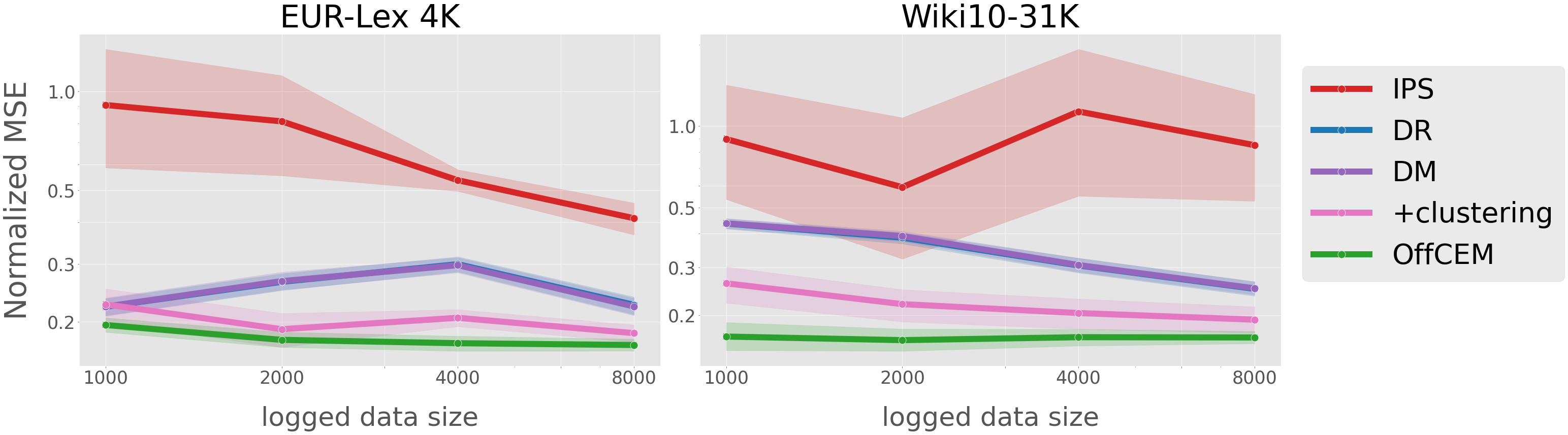}
    \end{center}
    \vspace{-3mm}
    \caption{Comparison of the estimators' MSE with varying logged data sizes on the real-world extreme classification datasets.}
    \label{fig:real}
\end{minipage}
\end{figure*}
\subsection{Real-World Data} \label{sec:real}
To assess the real-world applicability of OffCEM, we now evaluate OffCEM on extreme classification data. Specifically, we use EUR-Lex 4K and Wiki10-31K from the Extreme Classification Repository~\citep{bhatia2016extreme}, which were used in previous work on off-policy learning in large action spaces~\citep{lopez2021learning}. Table~\ref{tab:data_stats} provides the statistic of these datasets where we can see that both datasets contain several thousands of labels (actions).

Following prior work~\citep{dudik2014doubly,wang2017optimal,su2019cab,su2020doubly}, we convert the extreme classification datasets with $L$ labels into contextual bandit datasets with the same number of actions. We consider stochastic rewards where we first define the base reward function as follows.
\begin{align*}
    \tilde{q}(x, a)=
    \left\{\begin{array}{ll}1-\eta_a & \text {if $a$ is a positive label} \\ \eta_a - 1 & \text {otherwise}\end{array}\right.
\end{align*}
$\eta_a$ is a noise parameter sampled separately for each action $a$ from a uniform distribution with range $[0,0.2]$. Then, for each data, we sample the reward from a Bernoulli distribution with mean $q(x,a) = \sigma(\tilde{q}(x,a))$ where $\sigma(z) = (1 + \exp(-z))^{-1}$ is the sigmoid function. 

We define the logging policy $\pi_0$ by applying the softmax function to an estimated reward function $\hat{q}(x,a)$ similarly to Eq.~\eqref{eq:synthetic_logging} where we use $\beta=30$ and obtain $\hat{q}(x,a)$ by ridge regression with full-information labels. We also define the target policy $\pi$ following Eq.~\eqref{eq:synthetic_target} with $\epsilon=0.1$.

\paragraph{Results.}
We compare OffCEM against DM, IPS, and DR. We cannot include MIPS in the real-world experiment since there is no action embeddings available in the extreme classification datasets. Instead, we include ``+clustering" from the previous section as the best approximation of MIPS. Note that we obtain $\hat{q}(x,a)$ by simply regressing the observed reward as in Eq.~\eqref{eq:one_step_reg} by a neural network with 3 hidden layers and use it for DM and DR. For OffCEM, we perform the two-step regression from Section~\ref{sec:two-step_regression} to obtain $\hat{f}(x,a) = \hat{g}_{\psi}(x,\phi(a)) + \hat{h}_{\theta}(x,a)$ where $\hat{g}_{\psi}$ and $\hat{h}_{\theta}$ are parameterized by a neural network. We use uniform random action clusters for OffCEM with $|\calC| = 100$ as a heuristic.

Figure~\ref{fig:real} shows the MSEs of the estimators with varying logged data sizes averaged over 100 OPE simulations performed with different random seeds. The figure demonstrates that our OffCEM estimator outperforms all baselines on both datasets due to its substantially reduced variance against the baselines and its small bias. ``+clustering" also performs better than DM, IPS, and DR, but it produces larger bias and variance than OffCEM. The results suggest the real-world applicability of OffCEM even with heuristically generated and likely suboptimal action clusters, however, we can still expect a further improvement of OffCEM with a more refined clustering procedure.

\section{Conclusion and Future Work} \label{sec:conclusion} 
In this paper, we studied the problem of OPE for large action spaces and proposed the conjunct effect model, as well as its associated OffCEM estimator. OffCEM performs importance weighting over action clusters and deals with the remaining bias via model-based estimation. We characterize the bias and variance of the new estimator and show that it has superior statistical properties compared to IPS, DR, and MIPS. We also describe how to optimize the regression model to minimize the bias and variance of OffCEM, leveraging its local correctness condition via the two-step procedure. Extensive experiments on synthetic and real-world data demonstrate that OffCEM provides a substantial reduction in MSE particularly for large actions spaces. 

The findings in our paper raise several intriguing questions for future work. In particular, we relied on a heuristic procedure to find a clustering of the actions in the real-world experiment. While OffCEM performed better than the benchmarks even with heuristic clustering, we conjecture that there are more refined clustering procedures as concurrently explored by~\citet{peng2023offline}. For example, it would be valuable to consider an iterative procedure to optimize clustering and regression model simultaneously to satisfy local correctness better. It may also be possible to learn more general types of action representations to bring in further improvements beyond mere clustering. Moreover, this paper only considered the statistical problem of estimating the value of a fixed new policy, so it would be interesting to use our estimator to enable a more efficient off-policy learning in the presence of many actions.

\section*{Acknowledgements}
This research was supported in part by NSF Awards IIS-1901168 and IIS-2008139. Yuta Saito was supported by Funai Overseas Scholarship. All content represents the opinion of the authors, which is not necessarily shared or endorsed by their respective employers and/or sponsors. 

\bibliography{main.bbl}

\begin{thebibliography}{35}
\providecommand{\natexlab}[1]{#1}
\providecommand{\url}[1]{\texttt{#1}}
\expandafter\ifx\csname urlstyle\endcsname\relax
  \providecommand{\doi}[1]{doi: #1}\else
  \providecommand{\doi}{doi: \begingroup \urlstyle{rm}\Url}\fi

\bibitem[Athey et~al.(2019)Athey, Chetty, Imbens, and Kang]{athey2019surrogate}
Athey, S., Chetty, R., Imbens, G.~W., and Kang, H.
\newblock The surrogate index: Combining short-term proxies to estimate
  long-term treatment effects more rapidly and precisely.
\newblock Technical report, National Bureau of Economic Research, 2019.

\bibitem[Athey et~al.(2020)Athey, Chetty, and Imbens]{athey2020combining}
Athey, S., Chetty, R., and Imbens, G.
\newblock Combining experimental and observational data to estimate treatment
  effects on long term outcomes.
\newblock \emph{arXiv preprint arXiv:2006.09676}, 2020.

\bibitem[Bhatia et~al.(2016)Bhatia, Dahiya, Jain, Kar, Mittal, Prabhu, and
  Varma]{bhatia2016extreme}
Bhatia, K., Dahiya, K., Jain, H., Kar, P., Mittal, A., Prabhu, Y., and Varma,
  M.
\newblock The extreme classification repository: Multi-label datasets and code,
  2016.
\newblock URL \url{http://manikvarma.org/downloads/XC/XMLRepository.html}.

\bibitem[Chen \& Ritzwoller(2021)Chen and Ritzwoller]{chen2021semiparametric}
Chen, J. and Ritzwoller, D.~M.
\newblock Semiparametric estimation of long-term treatment effects.
\newblock \emph{arXiv preprint arXiv:2107.14405}, 2021.

\bibitem[Dud{\'\i}k et~al.(2014)Dud{\'\i}k, Erhan, Langford, and
  Li]{dudik2014doubly}
Dud{\'\i}k, M., Erhan, D., Langford, J., and Li, L.
\newblock Doubly robust policy evaluation and optimization.
\newblock \emph{Statistical Science}, 29\penalty0 (4):\penalty0 485--511, 2014.

\bibitem[Farajtabar et~al.(2018)Farajtabar, Chow, and
  Ghavamzadeh]{farajtabar2018more}
Farajtabar, M., Chow, Y., and Ghavamzadeh, M.
\newblock More robust doubly robust off-policy evaluation.
\newblock In \emph{Proceedings of the 35th International Conference on Machine
  Learning}, volume~80, pp.\  1447--1456. PMLR, 2018.

\bibitem[Felicioni et~al.(2022)Felicioni, Ferrari~Dacrema, Restelli, and
  Cremonesi]{felicioni2022off}
Felicioni, N., Ferrari~Dacrema, M., Restelli, M., and Cremonesi, P.
\newblock Off-policy evaluation with deficient support using side information.
\newblock \emph{Advances in Neural Information Processing Systems}, 35, 2022.

\bibitem[Jiang \& Li(2016)Jiang and Li]{jiang2016doubly}
Jiang, N. and Li, L.
\newblock Doubly robust off-policy value evaluation for reinforcement learning.
\newblock In \emph{Proceedings of the 33rd International Conference on Machine
  Learning}, volume~48, pp.\  652--661. PMLR, 2016.

\bibitem[Kallus \& Uehara(2020)Kallus and Uehara]{kallus2020double}
Kallus, N. and Uehara, M.
\newblock Double reinforcement learning for efficient off-policy evaluation in
  markov decision processes.
\newblock \emph{J. Mach. Learn. Res.}, 21:\penalty0 167--1, 2020.

\bibitem[Kallus et~al.(2021)Kallus, Saito, and Uehara]{kallus2020optimal}
Kallus, N., Saito, Y., and Uehara, M.
\newblock Optimal off-policy evaluation from multiple logging policies.
\newblock In \emph{Proceedings of the 38th International Conference on Machine
  Learning}, volume 139, pp.\  5247--5256. PMLR, 2021.

\bibitem[Lee et~al.(2022)Lee, Arbour, and Theocharous]{lee2022off}
Lee, J.~J., Arbour, D., and Theocharous, G.
\newblock Off-policy evaluation in embedded spaces.
\newblock \emph{arXiv preprint arXiv:2203.02807}, 2022.

\bibitem[Liu et~al.(2018)Liu, Li, Tang, and Zhou]{liu2018breaking}
Liu, Q., Li, L., Tang, Z., and Zhou, D.
\newblock Breaking the curse of horizon: infinite-horizon off-policy
  estimation.
\newblock In \emph{Proceedings of the 32nd International Conference on Neural
  Information Processing Systems}, pp.\  5361--5371, 2018.

\bibitem[Liu et~al.(2020)Liu, Bacon, and Brunskill]{liu2020understanding}
Liu, Y., Bacon, P.-L., and Brunskill, E.
\newblock Understanding the curse of horizon in off-policy evaluation via
  conditional importance sampling.
\newblock In \emph{International Conference on Machine Learning}, pp.\
  6184--6193. PMLR, 2020.

\bibitem[Lopez et~al.(2021)Lopez, Dhillon, and Jordan]{lopez2021learning}
Lopez, R., Dhillon, I.~S., and Jordan, M.~I.
\newblock Learning from extreme bandit feedback.
\newblock \emph{Proc. Association for the Advancement of Artificial
  Intelligence}, 2021.

\bibitem[Metelli et~al.(2021)Metelli, Russo, and
  Restelli]{metelli2021subgaussian}
Metelli, A.~M., Russo, A., and Restelli, M.
\newblock Subgaussian and differentiable importance sampling for off-policy
  evaluation and learning.
\newblock \emph{Advances in Neural Information Processing Systems}, 34, 2021.

\bibitem[Narita et~al.(2019)Narita, Yasui, and Yata]{narita2019efficient}
Narita, Y., Yasui, S., and Yata, K.
\newblock Efficient counterfactual learning from bandit feedback.
\newblock In \emph{Proceedings of the AAAI Conference on Artificial
  Intelligence}, volume~33, pp.\  4634--4641, 2019.

\bibitem[Newey \& Robins(2018)Newey and Robins]{newey2018crossfitting}
Newey, W.~K. and Robins, J.~R.
\newblock Cross-fitting and fast remainder rates for semiparametric estimation.
\newblock \emph{arXiv preprint arXiv:1801.09138}, 2018.

\bibitem[Pedregosa et~al.(2011)Pedregosa, Varoquaux, Gramfort, Michel, Thirion,
  Grisel, Blondel, Prettenhofer, Weiss, Dubourg, Vanderplas, Passos,
  Cournapeau, Brucher, Perrot, and {{\'E}}douard
  Duchesnay]{pedregosa2011scikit}
Pedregosa, F., Varoquaux, G., Gramfort, A., Michel, V., Thirion, B., Grisel,
  O., Blondel, M., Prettenhofer, P., Weiss, R., Dubourg, V., Vanderplas, J.,
  Passos, A., Cournapeau, D., Brucher, M., Perrot, M., and {{\'E}}douard
  Duchesnay.
\newblock Scikit-learn: Machine learning in python.
\newblock \emph{Journal of Machine Learning Research}, 12:\penalty0 2825--2830,
  2011.

\bibitem[Peng et~al.(2023)Peng, Zou, Liu, Li, Jiang, Pei, and
  Cui]{peng2023offline}
Peng, J., Zou, H., Liu, J., Li, S., Jiang, Y., Pei, J., and Cui, P.
\newblock Offline policy evaluation in large action spaces via outcome-oriented
  action grouping.
\newblock In \emph{Proceedings of the ACM Web Conference 2023}, pp.\
  1220--1230, 2023.

\bibitem[Sachdeva et~al.(2020)Sachdeva, Su, and Joachims]{sachdeva2020off}
Sachdeva, N., Su, Y., and Joachims, T.
\newblock Off-policy bandits with deficient support.
\newblock In \emph{Proceedings of the 26th ACM SIGKDD International Conference
  on Knowledge Discovery and Data Mining}, pp.\  965--975, 2020.

\bibitem[Saito \& Joachims(2021)Saito and Joachims]{saito2021counterfactual}
Saito, Y. and Joachims, T.
\newblock Counterfactual learning and evaluation for recommender systems:
  Foundations, implementations, and recent advances.
\newblock In \emph{Proceedings of the 15th ACM Conference on Recommender
  Systems}, pp.\  828–830, 2021.

\bibitem[Saito \& Joachims(2022)Saito and Joachims]{saito2022off}
Saito, Y. and Joachims, T.
\newblock Off-policy evaluation for large action spaces via embeddings.
\newblock In \emph{International Conference on Machine Learning}, pp.\
  19089--19122. PMLR, 2022.

\bibitem[Saito et~al.(2021{\natexlab{a}})Saito, Aihara, Matsutani, and
  Narita]{saito2021open}
Saito, Y., Aihara, S., Matsutani, M., and Narita, Y.
\newblock Open bandit dataset and pipeline: Towards realistic and reproducible
  off-policy evaluation.
\newblock In \emph{Thirty-fifth Conference on Neural Information Processing
  Systems Datasets and Benchmarks Track}, 2021{\natexlab{a}}.

\bibitem[Saito et~al.(2021{\natexlab{b}})Saito, Udagawa, Kiyohara, Mogi,
  Narita, and Tateno]{saito2021evaluating}
Saito, Y., Udagawa, T., Kiyohara, H., Mogi, K., Narita, Y., and Tateno, K.
\newblock Evaluating the robustness of off-policy evaluation.
\newblock In \emph{Proceedings of the 15th ACM Conference on Recommender
  Systems}, pp.\  114–123, 2021{\natexlab{b}}.

\bibitem[Su et~al.(2019)Su, Wang, Santacatterina, and Joachims]{su2019cab}
Su, Y., Wang, L., Santacatterina, M., and Joachims, T.
\newblock Cab: Continuous adaptive blending for policy evaluation and learning.
\newblock In \emph{International Conference on Machine Learning}, volume~84,
  pp.\  6005--6014, 2019.

\bibitem[Su et~al.(2020{\natexlab{a}})Su, Dimakopoulou, Krishnamurthy, and
  Dud{\'\i}k]{su2020doubly}
Su, Y., Dimakopoulou, M., Krishnamurthy, A., and Dud{\'\i}k, M.
\newblock Doubly robust off-policy evaluation with shrinkage.
\newblock In \emph{Proceedings of the 37th International Conference on Machine
  Learning}, volume 119, pp.\  9167--9176. PMLR, 2020{\natexlab{a}}.

\bibitem[Su et~al.(2020{\natexlab{b}})Su, Srinath, and
  Krishnamurthy]{su2020adaptive}
Su, Y., Srinath, P., and Krishnamurthy, A.
\newblock Adaptive estimator selection for off-policy evaluation.
\newblock In \emph{International Conference on Machine Learning}, pp.\
  9196--9205. PMLR, 2020{\natexlab{b}}.

\bibitem[Swaminathan \& Joachims(2015{\natexlab{a}})Swaminathan and
  Joachims]{swaminathan2015batch}
Swaminathan, A. and Joachims, T.
\newblock Batch learning from logged bandit feedback through counterfactual
  risk minimization.
\newblock \emph{The Journal of Machine Learning Research}, 16\penalty0
  (1):\penalty0 1731--1755, 2015{\natexlab{a}}.

\bibitem[Swaminathan \& Joachims(2015{\natexlab{b}})Swaminathan and
  Joachims]{swaminathan2015counterfactual}
Swaminathan, A. and Joachims, T.
\newblock Counterfactual risk minimization: Learning from logged bandit
  feedback.
\newblock In \emph{International Conference on Machine Learning}, pp.\
  814--823. PMLR, 2015{\natexlab{b}}.

\bibitem[Swaminathan \& Joachims(2015{\natexlab{c}})Swaminathan and
  Joachims]{swaminathan2015self}
Swaminathan, A. and Joachims, T.
\newblock The self-normalized estimator for counterfactual learning.
\newblock \emph{Advances in Neural Information Processing Systems}, 28,
  2015{\natexlab{c}}.

\bibitem[Thomas \& Brunskill(2016)Thomas and Brunskill]{thomas2016data}
Thomas, P. and Brunskill, E.
\newblock Data-efficient off-policy policy evaluation for reinforcement
  learning.
\newblock In \emph{Proceedings of the 33rd International Conference on Machine
  Learning}, volume~48, pp.\  2139--2148. PMLR, 2016.

\bibitem[Udagawa et~al.(2023)Udagawa, Kiyohara, Narita, Saito, and
  Tateno]{udagawa2023policy}
Udagawa, T., Kiyohara, H., Narita, Y., Saito, Y., and Tateno, K.
\newblock Policy-adaptive estimator selection for off-policy evaluation.
\newblock In \emph{Proceedings of the AAAI Conference on Artificial
  Intelligence}, volume~36, 2023.

\bibitem[Voloshin et~al.(2019)Voloshin, Le, Jiang, and
  Yue]{voloshin2019empirical}
Voloshin, C., Le, H.~M., Jiang, N., and Yue, Y.
\newblock Empirical study of off-policy policy evaluation for reinforcement
  learning.
\newblock \emph{arXiv preprint arXiv:1911.06854}, 2019.

\bibitem[Wang et~al.(2017)Wang, Agarwal, and Dud{\i}k]{wang2017optimal}
Wang, Y.-X., Agarwal, A., and Dud{\i}k, M.
\newblock Optimal and adaptive off-policy evaluation in contextual bandits.
\newblock In \emph{International Conference on Machine Learning}, pp.\
  3589--3597. PMLR, 2017.

\bibitem[Xie et~al.(2019)Xie, Ma, and Wang]{xie2019towards}
Xie, T., Ma, Y., and Wang, Y.-X.
\newblock Towards optimal off-policy evaluation for reinforcement learning with
  marginalized importance sampling.
\newblock In \emph{Advances in Neural Information Processing Systems}, pp.\
  9665--9675, 2019.

\end{thebibliography}
\bibliographystyle{icml2023}

\newpage
\appendix
\onecolumn

\section{Related Work} \label{app:related}
Off-policy evaluation of fixed decision-making policies is of great practical relevance, providing a safe and cost-effective alternative for online A/B testing~\citep{saito2021counterfactual}, and thus has gained particular attention in both contextual bandits~\citep{dudik2014doubly,wang2017optimal,liu2018breaking,farajtabar2018more,su2019cab,su2020doubly,kallus2020optimal,metelli2021subgaussian} and reinforcement learning (RL)~\citep{jiang2016doubly,thomas2016data,xie2019towards,kallus2020double,liu2020understanding}.
There are three benchmark estimators in this area. The first estimator is DM, which estimates the expected reward function by some off-the-self supervised machine learning methods and then estimates the value of new policies based on the estimated rewards. DM has a lower variance than IPS, but it is susceptible to misspecification of the expected reward function, which is often the case in highly complex real-world data~\citep{farajtabar2018more,voloshin2019empirical}. The second benchmark estimator is IPS, which applies importance weighting to the observed rewards to estimate the policy value. Under some identification assumptions such as common support and unconfoundedness, IPS enables unbiased and consistent OPE. However, a drawback is that it can produce excessive bias and variance in large action spaces. First, it can have a high bias when the logging policy fails to satisfy the common support condition, which is expected when there are many actions~\citep{sachdeva2020off}. There is also the critical variance issue, as the importance weights are likely to be extremely large in the presence of many actions. The weight clipping~\cite{swaminathan2015counterfactual,su2019cab,su2020doubly} and normalization~\citep{swaminathan2015self} might be applied to deal with the variance issue, but these simple techniques often produce additional bias. Thus, DR has gained particular attention as a hybrid of DM and IPS, which can achieve a lower bias than DM and a lower variance than IPS~\citep{dudik2014doubly,wang2017optimal,farajtabar2018more,su2020doubly}. Though there are a number of extensions of DR both in contextual bandits~\citep{dudik2014doubly,wang2017optimal,farajtabar2018more,su2020doubly,metelli2021subgaussian} and RL~\citep{jiang2016doubly,thomas2016data,kallus2020double}, it may still degrade drastically due to extreme variance in large action spaces (as shown in Eq.~\eqref{eq:dr_variance}). This is because DR relies on the vanilla importance weight defined with respect to the distributions over large action spaces, causing the critical variance issue as in IPS. To overcome this fundamental variance issue of the typical OPE estimators for large action spaces, \citet{saito2022off} recently propose a new framework and estimator called MIPS, which utilize some prior information about the actions (i.e., action embeddings or action features) that are widely available in practice and provide structure in the action space. More specifically, MIPS is based on the marginal importance weight, which is defined with respect to the marginal distributions of the action embeddings induced by the target and logging policies. By doing so, it was shown that MIPS has increasingly lower variance than IPS in larger action spaces while remaining unbiased under the no direct effect assumption, which requires that the given action embeddings should be informative enough to be able to mediate every causal effect of the actions on the rewards (Assumption~\ref{assumption:no_direct_effect}). A similar assumption is also often utilized when performing causal inference of long-term outcomes via short-term surrogates~\citep{athey2019surrogate,athey2020combining,chen2021semiparametric} and to deal with the deficient support problem in OPE~\citep{felicioni2022off}. Moreover, \citet{lee2022off} propose a method based on normalizing flow to estimate the marginal importance weight when the logging and target policies are unknown. Note that, in order to improve the MSE of MIPS, we can perform some data-driven action feature selection in a way that minimizes the MSE of the resulting estimator based on existing estimator selection methods for OPE~\citep{su2020adaptive,udagawa2023policy}. However, a caveat is that MIPS may still have a very large variance similarly to IPS when the given action embeddings are high-dimensional and fined-grained. Moreover, it may produce a large bias if the no direct effect is violated and there is much direct effect of the actions that is not captured by the action embeddings. This bias issue is particularly expected when we perform action feature selection of high-dimensional and granular action embeddings aiming for variance reduction. \textbf{Therefore, there remains a critical bias-variance dilemma regarding MIPS -- high-dimensional action embeddings should be avoided to guarantee sufficient variance reduction, however, naively doing so via action feature selection might produce large bias by violating no direct effect}. The goal of our work is thus to overcome this bias-variance dilemma of MIPS by generalizing the formulation and proposing a refined estimator. Specifically, rather than making no direct effect, which is often stringent, we decompose the expected reward function into what we call the cluster effect and residual effect. We then apply model-free estimation based on cluster importance weight to estimate the cluster effect with no bias and apply model-based estimation based on the pairwise regression procedure to estimate the residual effect with small variance rather than applying marginal importance weight to both terms as in MIPS, possibly producing excessive variance. Our OffCEM estimator thus achieves much smaller variance than IPS, DR, and MIPS in the presence of many actions or high-dimensional action embeddings, while often reducing the bias of MIPS without ignoring the residual effect.

\newpage
\begin{table*}[h]
\begin{minipage}{\textwidth}
    \caption{Examples of locally correct regression models} \label{tab:example_regression_models}
    \vspace{0.1in}
  \begin{minipage}[t]{.33\textwidth}
    \centering
    \scalebox{0.95}{
    \begin{tabular}{c|cc|cc}
        \toprule
        $a$ & $a_0$ & $a_1$ & $a_2$ & $a_3$ \\ \midrule \midrule
        $\phi(x_0,a)$ & \multicolumn{2}{c|}{0} & \multicolumn{2}{c}{1}  \\
        $q(x_0,a)$ & 4 & 1 & 3 & 2  \\
        $\hat{f}_1(x_0,a)$ & 3 & 0 & 1 & 0  \\ \midrule
        $\Delta_{\cdot,\cdot}(x_0,a,b) $ & \multicolumn{2}{c|}{3} & \multicolumn{2}{c}{1}  \\ 
        \bottomrule
    \end{tabular}}
  \end{minipage}
  \hfill
  \begin{minipage}[t]{.33\textwidth}
    \centering
    \scalebox{0.95}{
    \begin{tabular}{c|cc|cc}
        \toprule
        $a$ & $a_0$ & $a_1$ & $a_2$ & $a_3$ \\ \midrule \midrule
        $\phi(x_0,a)$ & \multicolumn{2}{c|}{0} & \multicolumn{2}{c}{1}  \\
        $q(x_0,a)$ & 4 & 1 & 3 & 2  \\
        $\hat{f}_2(x_0,a)$ & 50 & 47 & -30 & -31  \\ \midrule
        $\Delta_{\cdot,\cdot}(x_0,a,b) $ & \multicolumn{2}{c|}{3} & \multicolumn{2}{c}{1}  \\ 
        \bottomrule
    \end{tabular}}
  \end{minipage}
  \hfill
  \begin{minipage}[t]{.33\textwidth}
    \centering
    \scalebox{0.95}{
    \begin{tabular}{c|cc|cc}
        \toprule
        $a$ & $a_0$ & $a_1$ & $a_2$ & $a_3$ \\ \midrule \midrule
        $\phi(x_0,a)$ & \multicolumn{2}{c|}{0} & \multicolumn{2}{c}{1}  \\
        $q(x_0,a)$ & 4 & 1 & 3 & 2  \\
        $\hat{f}_3(x_0,a)$ & 4 & 1 & 3 & 2  \\ \midrule
        $\Delta_{\cdot,\cdot}(x_0,a,b) $ & \multicolumn{2}{c|}{3} & \multicolumn{2}{c}{1}  \\ 
        \bottomrule
    \end{tabular}}
  \end{minipage}
\end{minipage}
\end{table*}
\section{Examples: Locally Correct Regression Models} \label{app:local_correctness}

This section provides some examples of regression model $\hat{f}$ that satisfy Assumption~\ref{assumption:local_correctness} (local correctness). Suppose that there is only a single context $\calX = \{x_0\}$ and four actions $\calA = \{a_0, a_1, a_2, a_3\}$. The expected reward function $q(x,a)$ and clustering function $\phi(x,a)$ are given as follows.
\begin{align*}
    q(x_0,a_0) = 4, \; q(x_0,a_1) = 1, \; q(x_0,a_2) = 3, \; q(x_0,a_3) = 2,\\
    \phi(x_0,a_0) = 0, \; \phi(x_0,a_1) = 0, \; \phi(x_0,a_2) = 1, \; \phi(x_0,a_3) = 1.
\end{align*}
Then, Table~\ref{tab:example_regression_models} provides three locally correct regression models ($\hat{f}_1$ to $\hat{f}_3$). 
More specifically, these example models succeed in preserving the relative value difference of the actions within each action cluster. In fact, we can see that $\Delta_q(x_0,a_0,a_1) = \Delta_{\hat{f}_1}(x_0,a_0,a_1) = \Delta_{\hat{f}_2}(x_0,a_0,a_1) = \Delta_{\hat{f}_3}(x_0,a_0,a_1) = 3$ and $\Delta_q(x_0,a_2,a_3) = \Delta_{\hat{f}_1}(x_0,a_2,a_3) = \Delta_{\hat{f}_2}(x_0,a_2,a_3) = \Delta_{\hat{f}_3}(x_0,a_2,a_3) =  1$ where $\phi(x_0,a_0)=\phi(x_0,a_1)$ and $\phi(x_0,a_2)=\phi(x_0,a_3)$.

\begin{figure*}[h]
\centering
\begin{minipage}{\hsize}
    \begin{center}
        \includegraphics[clip, width=16.5cm]{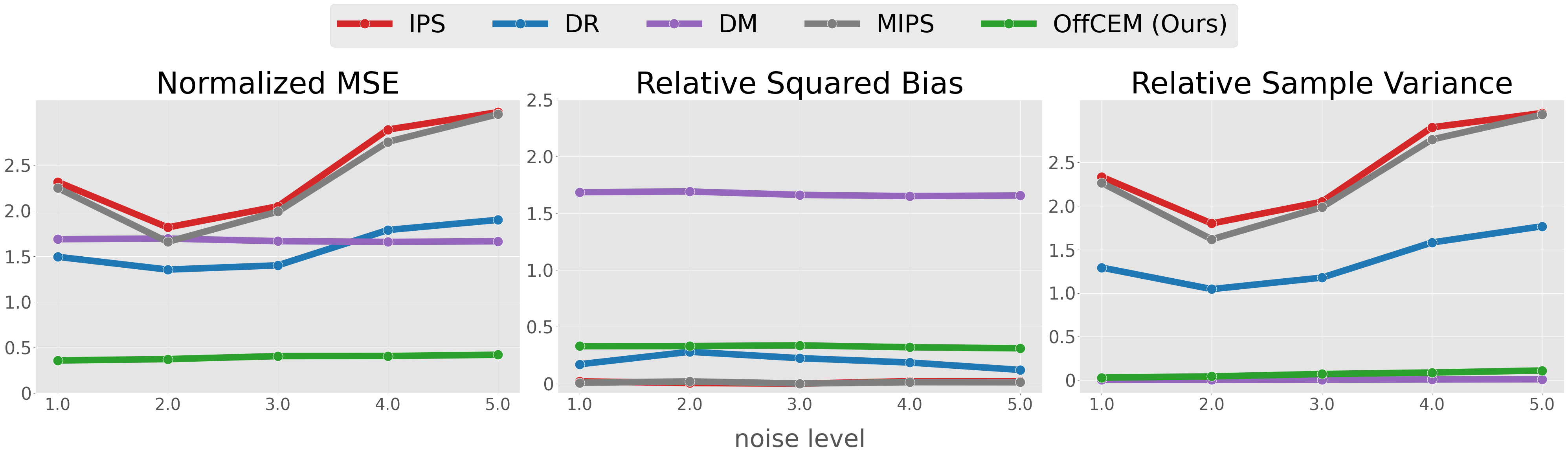}
    \end{center}
    \vspace{-2mm}
    \caption{Comparison of the estimators' MSE (normalized by the true value $V(\pi)$) with \textbf{varying noise levels} ($\sigma$).}
    \label{fig:varying_noise}
\end{minipage}
\\ \vspace{15mm}
\begin{minipage}{\hsize}
    \begin{center}
        \includegraphics[clip, width=16.5cm]{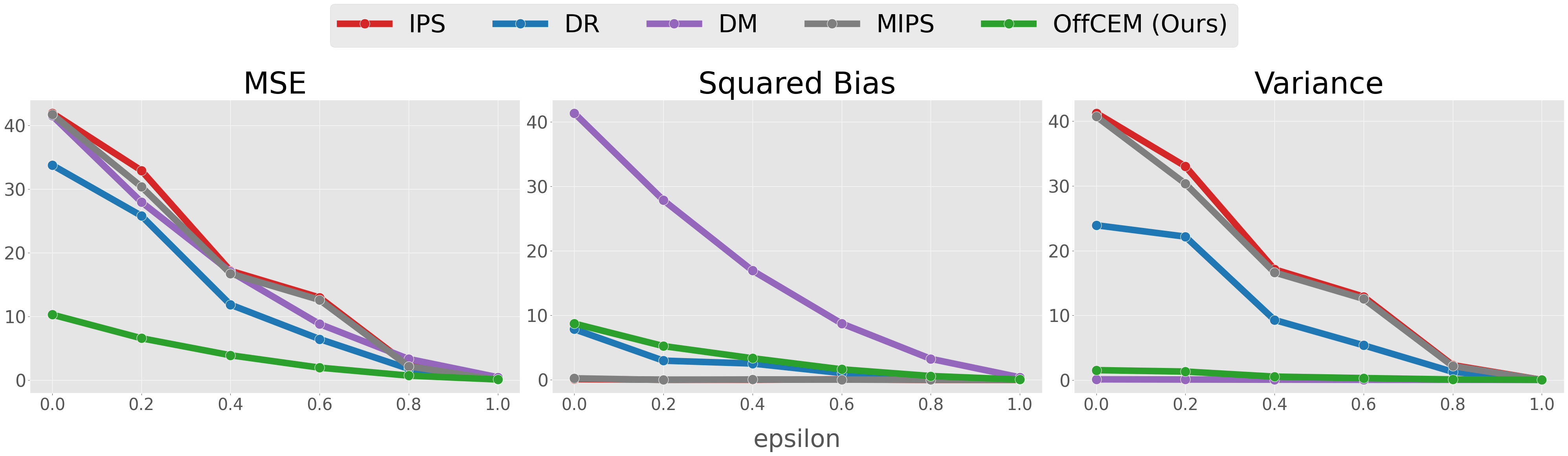}
    \end{center}
    \vspace{-2mm}
    \caption{Comparison of the estimators' MSE with \textbf{varying target policies} ($\epsilon$).}
    \label{fig:varying_eps}
\end{minipage}
\\ \vspace{15mm}
\begin{minipage}{\hsize}
    \begin{center}
        \includegraphics[clip, width=16.5cm]{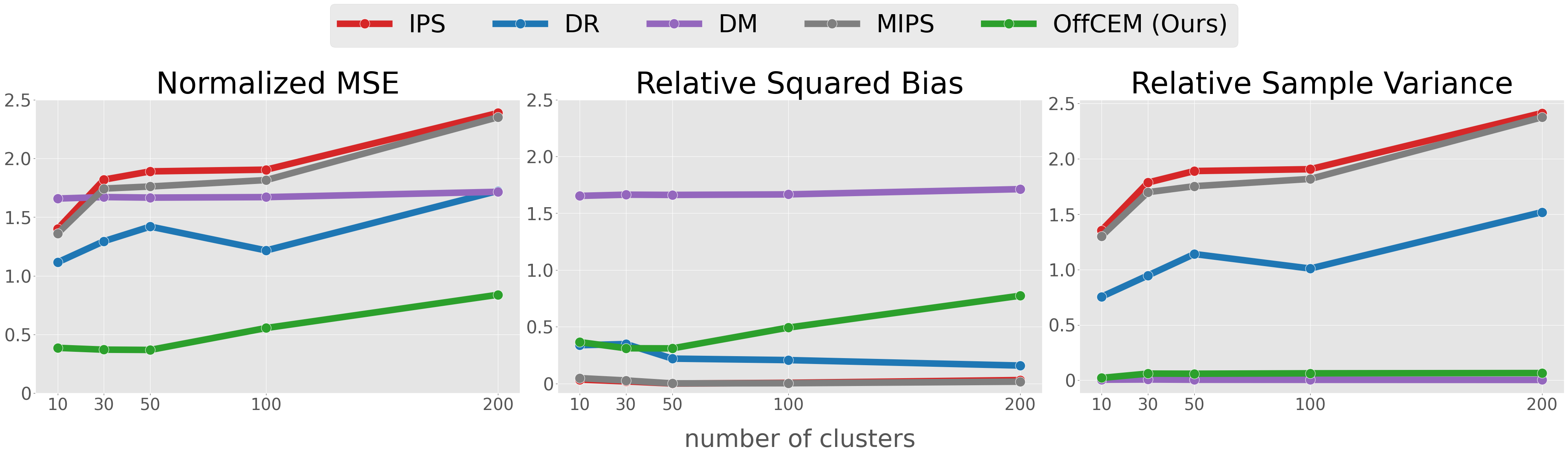}
    \end{center}
    \vspace{-2mm}
    \caption{Comparison of the estimators' MSE (normalized by the true value $V(\pi)$) with \textbf{varying numbers of clusters} ($|\calC|$).}
    \label{fig:varying_n_clusters}
\end{minipage}
\vskip 0.1in
\raggedright
\fontsize{10pt}{10pt}\selectfont \textit{Note}:
We set $|\calA|=1,000$, $|\calC|=50$, $\epsilon=0.2$, $\beta=-0.1$, and $\sigma=3.0$ as default experiment parameters.
The results are averaged over 300 different sets of synthetic logged data replicated with different random seeds.
The shaded regions in the MSE plots represent the 95\% confidence intervals estimated with bootstrap.
\end{figure*}

\begin{figure*}[h]
\centering
\begin{minipage}{\hsize}
    \begin{center}
        \includegraphics[clip, width=16.5cm]{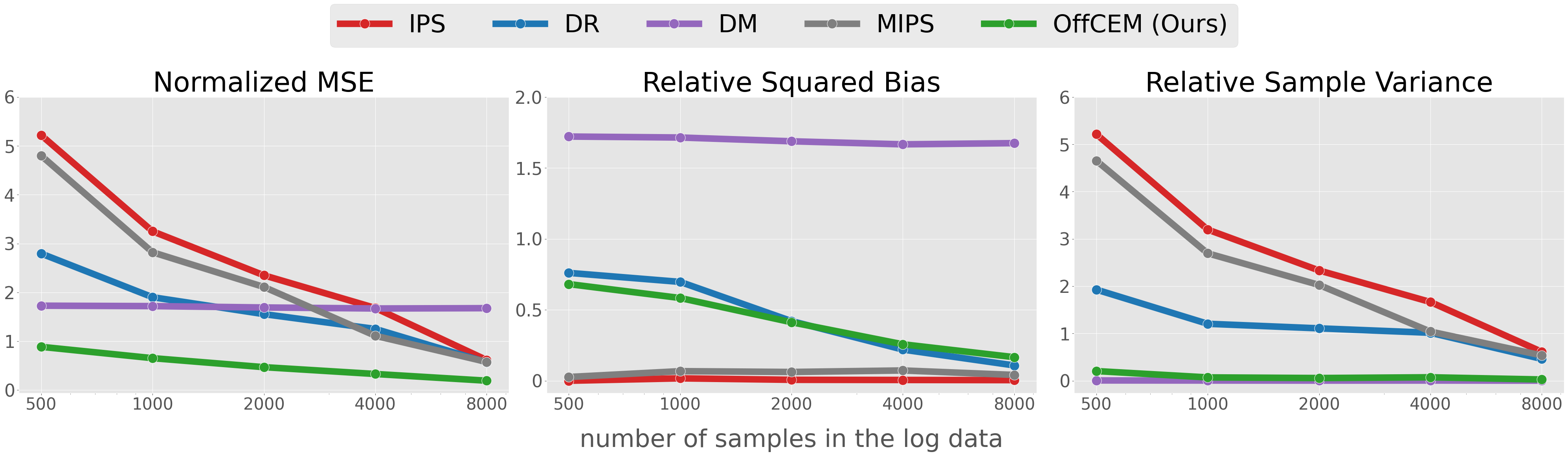}
    \end{center}
    \vspace{-2mm}
    \caption{MSE, Squared Bias, and Variance with \textbf{varying sample sizes} ($n$).}
    \label{fig:varying_n_val}
\end{minipage}
\\ \vspace{1in}
\begin{minipage}{\hsize}
    \begin{center}
        \includegraphics[clip, width=16.5cm]{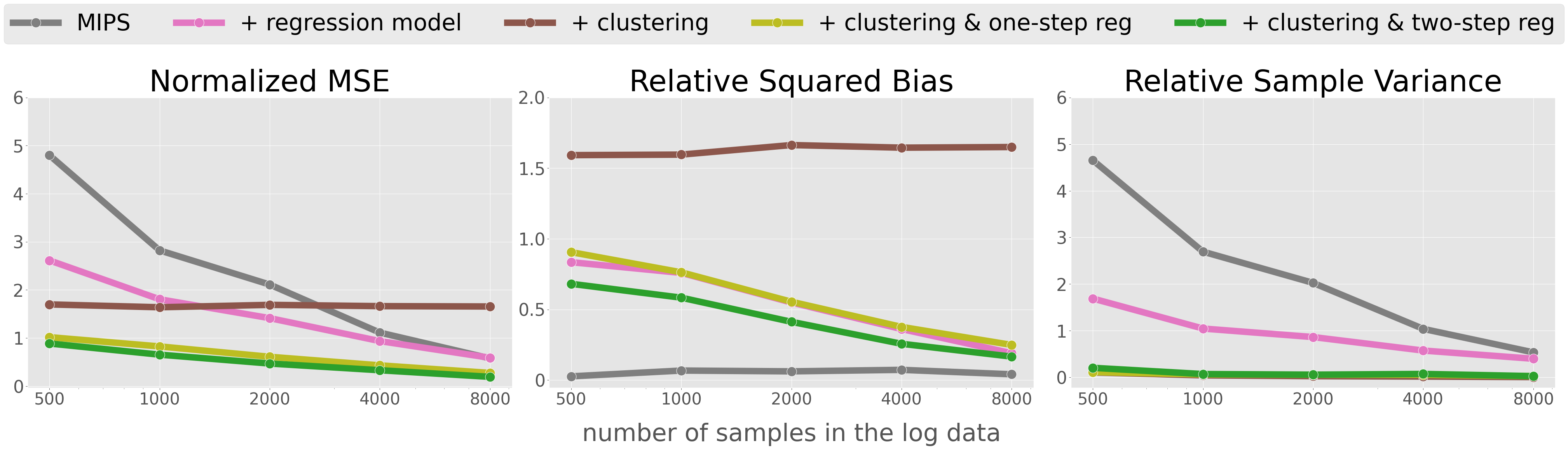}
    \end{center}
    \vspace{-2mm}
    \caption{MSE, Squared Bias, and Variance (ablation) with \textbf{varying sample sizes} ($n$).}
    \label{fig:varying_n_val_ablation}
\end{minipage}
\vskip 0.1in
\raggedright
\fontsize{10pt}{10pt}\selectfont \textit{Note}:
We set $|\calA|=1,000$, $|\calC|=50$, $\epsilon=0.2$, $\beta=-0.1$, and $\sigma=3.0$ as default experiment parameters.
The results are averaged over 300 different sets of synthetic logged data replicated with different random seeds.
The shaded regions in the MSE plots represent the 95\% confidence intervals estimated with bootstrap.
\end{figure*}

\begin{figure*}[h]
\centering
\begin{minipage}{\hsize}
    \begin{center}
        \includegraphics[clip, width=16.5cm]{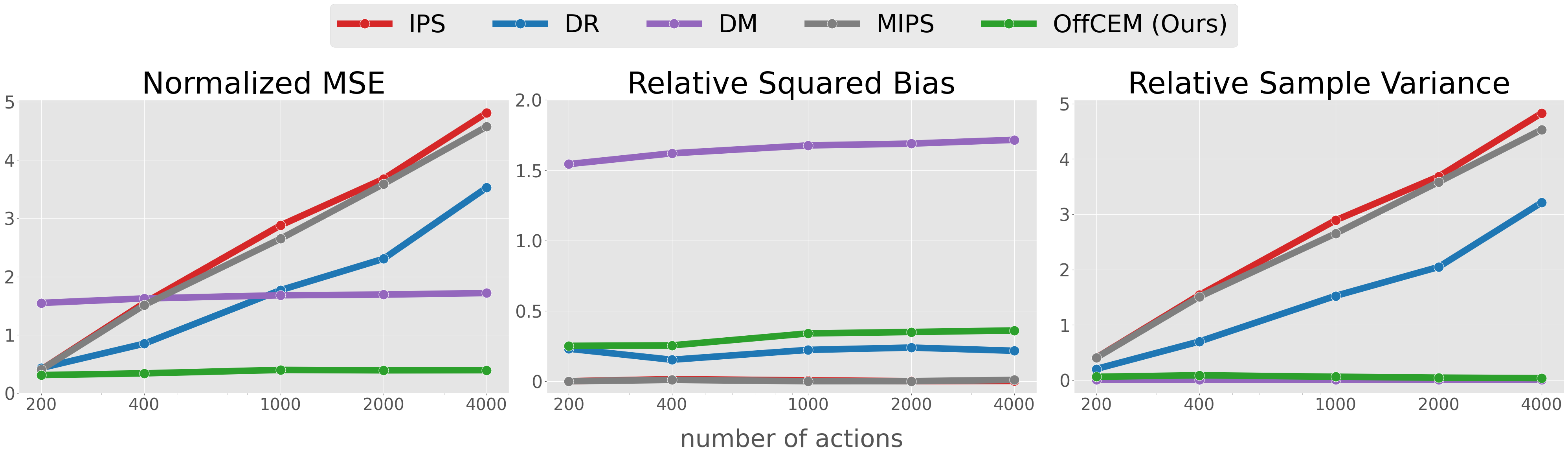}
    \end{center}
    \vspace{-2mm}
    \caption{MSE, Squared Bias, and Variance with \textbf{varying numbers of actions} ($|\calA|$).}
    \label{fig:varying_n_actions}
\end{minipage}
\\ \vspace{1in}
\begin{minipage}{\hsize}
    \begin{center}
        \includegraphics[clip, width=16.5cm]{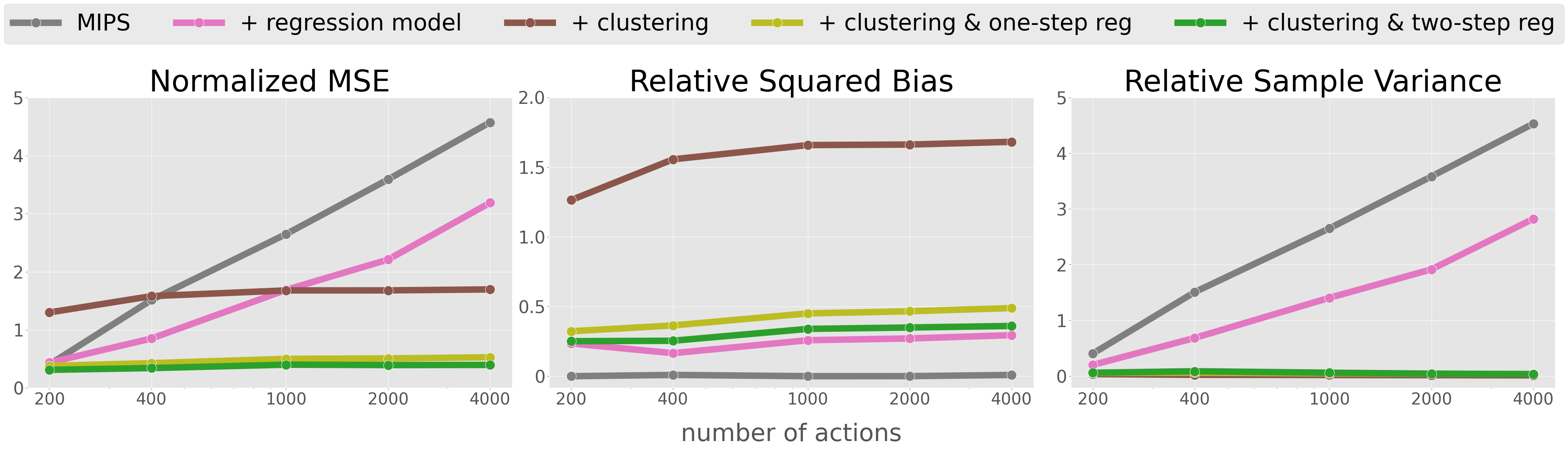}
    \end{center}
    \vspace{-2mm}
    \caption{MSE, Squared Bias, and Variance (ablation) with \textbf{varying numbers of actions} ($|\calA|$).}
    \label{fig:varying_actions_ablation}
\end{minipage}
\vskip 0.1in
\raggedright
\fontsize{10pt}{10pt}\selectfont \textit{Note}:
We set $n=3,000$, $|\calC|=50$, $\epsilon=0.2$, $\beta=-0.1$, and $\sigma=3.0$ as default experiment parameters.
The results are averaged over 300 different sets of synthetic logged data replicated with different random seeds.
The shaded regions in the MSE plots represent the 95\% confidence intervals estimated with bootstrap.
\end{figure*}

\begin{figure*}[h]
\centering
\begin{minipage}{\hsize}
    \begin{center}
        \includegraphics[clip, width=16.5cm]{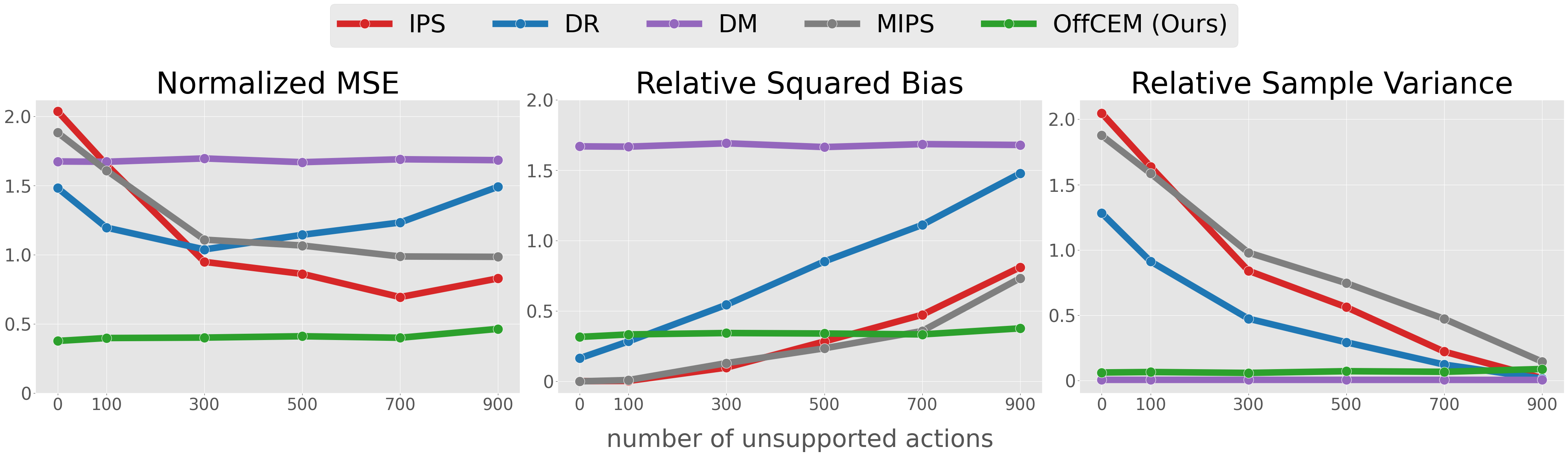}
    \end{center}
    \vspace{-2mm}
    \caption{MSE, Squared Bias, and Variance with \textbf{varying numbers of unsupported actions}.}
    \label{fig:varying_n_def_actions}
\end{minipage}
\\ \vspace{1in}
\begin{minipage}{\hsize}
    \begin{center}
        \includegraphics[clip, width=16.5cm]{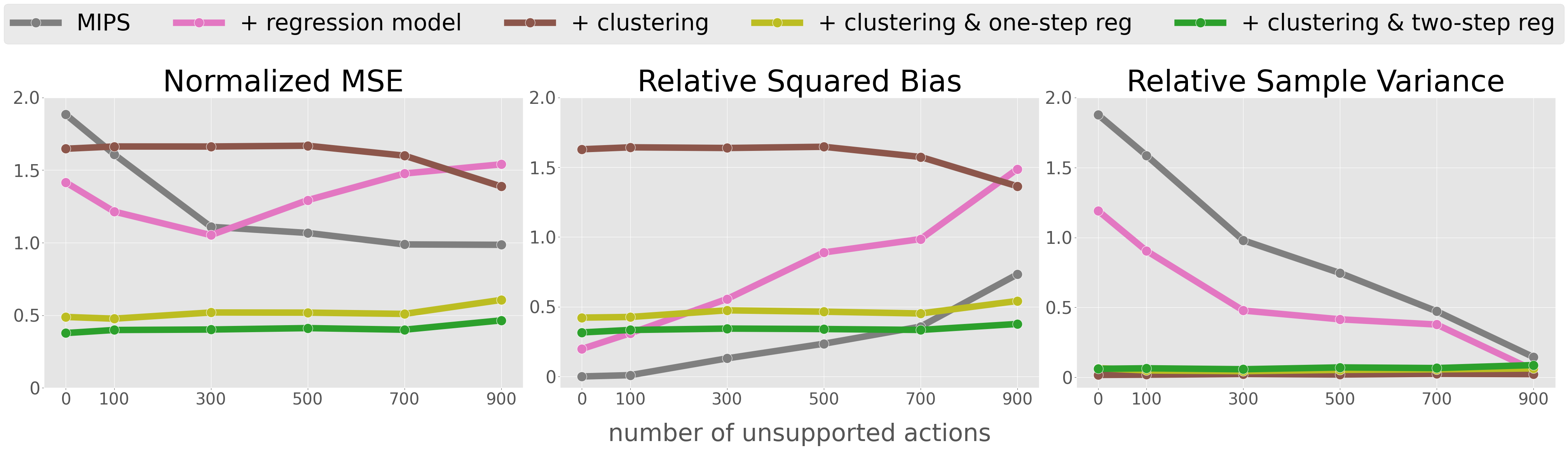}
    \end{center}
    \vspace{-2mm}
    \caption{MSE, Squared Bias, and Variance (ablation) with \textbf{varying numbers of unsupported actions}.}
    \label{fig:varying_n_def_actions_ablation}
\end{minipage}
\vskip 0.1in
\raggedright
\fontsize{10pt}{10pt}\selectfont \textit{Note}:
We set $n=3,000$, $|\calA|=1,000$, $\epsilon=0.2$, $\beta=-0.1$, and $\sigma=3.0$ as default experiment parameters.
The results are averaged over 300 different sets of synthetic logged data replicated with different random seeds.
The shaded regions in the MSE plots represent the 95\% confidence intervals estimated with bootstrap.
\end{figure*}

\section{Omitted Proofs} \label{app:proof}
First, we generalize the formulation in the main content and will be based on action clustering given deterministic action embedding $e \in \calE$ and context $x$, i.e., $\phi: \calX \times \calE \rightarrow \calC$ to prove the theorems. Under this generalization, OffCEM is refined as
\begin{align}
    \offcem := \meanN \bigg\{ w(x_i,\phi(x_i, e_{a_i})) (r_i - \hat{f}(x_i,a_i,e_{a_i})) + \hat{f}(x_i,\pi) \bigg\}, \label{eq:general_OffCEM}
\end{align}
where $\hat{f}(x,\pi) := \mE_{\pi(a|x)}[\hat{f}(x,a,e_a)]$ and $$ w(x,\phi(x,e_a)) := \frac{\pi(\phi(x,e_a) \,|\, x)}{\pi_0(\phi(x,e_a) \,|\, x)} = \frac{\sum_{a' \in \calA} \mathbb{I}\{\phi(x,e_a) = \phi(x,e_{a'})\} \pi(a' \,|\, x)}{\sum_{a' \in \calA} \mathbb{I}\{\phi(x,e_a) = \phi(x,e_{a'})\} \pi_0(a' \,|\, x)}. $$

Moreover, under this setup, the local correctness assumption in Assumption~\ref{assumption:local_correctness} is generalized as follows.
\begin{assumption} (Generalized Local Correctness) \label{assumption:general_local_correctness}
A regression model $\hat{f}(\cdot,\cdot)$ and action clustering function $\phi(\cdot,\cdot)$ satisfy local correctness if the following holds true:
\begin{align*}
    & \Delta_q(x,a,b) = \Delta_{\hat{f}} (x,a,b),
\end{align*}
for all $x\in\calX$ and $a,b\in\calA$ with $\phi(x,e_a)=\phi(x,e_b)$, where $\Delta_q(x,a,b) := q(x,a,e_a) - q(x,b,e_b)$ is the difference in the expected rewards between the actions $a$ and $b$ given context $x$, which we call the \textit{relative value difference} of the actions. 

Note that, given the fact: $\Delta_q(x,a,b) = \Delta_{\hat{f}} (x,a,b) \Rightarrow \Delta_{q,\hat{f}}(x,a) = \Delta_{q,\hat{f}} (x,b)$, the local correctness assumption can equivalently be described as:

A regression model $\hat{f}(\cdot,\cdot)$ and action clustering function $\phi(\cdot,\cdot)$ satisfy local correctness if there exists a cluster value function $g: \calX \times \calC \rightarrow \mathbb{R}$ that satisfies the following:
\begin{align}
    \Delta_{q,\hat{f}} (x,a) = g(x,\phi(x,e_a)) \; \Big(\Rightarrow q(x,a,e_a) = g(x,\phi(x,e_a)) + \hat{f}(x,a,e_a) \Big), \label{eq:general_local_correctness}
\end{align}
for all $x\in\calX$ and $a\in\calA$ where $\Delta_{q,\hat{f}} (x,a,e) := q(x,a,e) - \hat{f}(x,a,e)$ is an estimation error of the regression model $\hat{f}$ given context $x$, action $a$, and embedding $e$. 
\end{assumption}
Note that it is straightforward to see that Assumption~\ref{assumption:general_local_correctness} is reduced to a simpler version used in the main text (Assumption~\ref{assumption:local_correctness}) when given is an action clustering function that depends directly on the action, i.e., $\phi: \calX \times \calA \rightarrow \calC$.

\subsection{Proof of Proposition~\ref{prop:unbiased}} \label{app:unbias}
\begin{proof}
We can calculate the expectation of OffCEM under Assumption~\ref{assumption:general_local_correctness} below. 
\begin{align*}
    & \mE_{\calD} [\offcem] \\
    & = \mE_{p(x)\pi_0(a|x)p(r|x,a,e_a)} \left[ w(x,\phi(x,e_a)) (r - \hat{f}(x,a,e_a) ) + \hat{f}(x,\pi) \right]  \\
    & = \mE_{p(x)\pi_0(a|x)} \left[ w(x,\phi(x,e_a)) (q(x,a,e_a) - \hat{f}(x,a,e) ) + \hat{f}(x,\pi) \right] \\
    & = \mE_{p(x)\pi_0(a|x)} \left[ w(x,\phi(x,e_a)) \Delta_{q,\hat{f}}(x,a,e_a) + \hat{f}(x,\pi) \right] \\
    & = \mE_{p(x)} \left[  \hat{f}(x,\pi) + \sumA \pi_0(a\,|\,x) \frac{\pi(\phi(x,e_a)\,|\,x)}{\pi_0(\phi(x,e_a)\,|\,x)} \Delta_{q,\hat{f}}(x,a,e_a)  \right] \\
    & = \mE_{p(x)} \left[  \hat{f}(x,\pi) + \sumA \pi_0(a\,|\,x) \frac{\pi(\phi(x,e_a)\,|\,x)}{\pi_0(\phi(x,e_a)\,|\,x)} g(x,\phi(x,e_a))  \right] \quad \because \text{Eq.~\eqref{eq:general_local_correctness}}  \\
    & = \mE_{p(x)} \left[  \hat{f}(x,\pi) + \sumA \pi_0(a\,|\,x) \sumC \frac{\pi(c\,|\,x)}{\pi_0(c\,|\,x)} g(x,c) \mathbb{I} \{\phi(x,e_a) = c\} \right]  \\
    & = \mE_{p(x)} \left[  \hat{f}(x,\pi) + \sumC \frac{\pi(c\,|\,x)}{\pi_0(c\,|\,x)} g(x,c) \sumA \pi_0(a\,|\,x) \mathbb{I} \{\phi(x,e_a) = c\} \right]  \\
    & = \mE_{p(x)} \left[  \hat{f}(x,\pi) + \sumC \frac{\pi(c\,|\,x)}{\pi_0(c\,|\,x)} g(x,c) \pi_0(c\,|\,x) \right] \quad \because  \pi_0(c\,|\,x) = \sumA \pi_0(a\,|\,x) \mathbb{I} \{\phi(x,e_a) = c\} \\
    & = \mE_{p(x)} \left[  \hat{f}(x,\pi) + \sumC \pi(c\,|\,x) g(x,c) \right] \\
    & = \mE_{p(x)} \left[  \hat{f}(x,\pi) + \sumC g(x,c) \sumA \pi(a\,|\,x) \mathbb{I} \{\phi(x,e_a) = c\} \right] \quad \because  \pi(c\,|\,x) = \sumA \pi(a\,|\,x) \mathbb{I} \{\phi(x,e_a) = c\} \\
    & = \mE_{p(x)} \left[  \hat{f}(x,\pi) + \sumA \pi(a\,|\,x) \sumC g(x,c) \mathbb{I} \{\phi(x,e_a) = c\} \right] \\
    & = \mE_{p(x)} \left[  \sumA \pi(a\,|\,x) \left\{ g(x,\phi(x,e_a)) + \hat{f}(x,a,e_a) \right\} \right] \\
    & = \mE_{p(x)\pi(a|x)} \left[  q(x,a,e_a) \right] \quad \because  \text{Eq.~\eqref{eq:general_local_correctness}}  \\
    & = V(\pi)
\end{align*}
and thus OffCEM is unbiased under Assumption~\ref{assumption:general_local_correctness}.
\end{proof}

\subsection{Proof of Theorem~\ref{thm:bias}} \label{app:bias}
\begin{proof}
We first derive the bias of the general form of OffCEM below.\footnote{Here, we work on a slightly more general case with stochastic action embeddings based on $p(e\,|\,x,a)$ as in \citet{saito2022off}.}
\begin{align}
    & \biasoffcem \notag \\
    & = \mE_{p(x)\pi_0(a|x)p(e|x,a)p(r|x,a,e)} [ w(x,\phi(x,e)) (r - \hat{f}(x,a,e)) + \hat{f}(x,\pi) ] - V(\pi) \notag \\
    & = \mE_{p(x)\pi_0(a|x)p(e|x,a)} [ w(x,\phi(x,e)) (q(x,a,e) - \hat{f}(x,a,e) ) + \hat{f}(x,\pi) ] - \mE_{p(x)\pi(a|x)p(e|x,a)}[q(x,a,e)] \notag 
\end{align}
\begin{align}
    & =  \mE_{p(x)} \left[ \sumA \pi_0(a\,|\,x) \sumE p(e\,|\,x,a) \Delta_{q,\hat{f}}(x,a,e) w(x,\phi(x,e)) \right] \notag \\
    & \quad + \mE_{p(x)} \left[ \sumA\sumE \pi(a,e\,|\,x) \hat{f}(x,a,e) \right] -\mE_{p(x)}\left[\sumA \pi(a\,|\,x) \sumE \frac{\pi_0\left(e\,|\,x\right) \pi_0(a\,|\,x,e)}{\pi_0(a\,|\,x)}  q(x,a,e)\right] \notag \\
    & = \mE_{p(x)} \left[\sumA \pi_0(a\,|\,x) \sumE \frac{\pi_0(e\,|\,x) \pi_0(a\,|\,x,e)}{\pi_0(a\,|\,x)} \Delta_{q,\hat{f}}(x,a,e) \sumC w(x,c) \mathbb{I} \{\phi(x,e) = c\} \right] \notag \\
    & \quad + \mE_{p(x)} \left[ \sumA\sumE \pi_0(a,e\,|\,x) \frac{\pi(a,e\,|\,x)}{\pi_0(a,e\,|\,x)} \hat{f}(x,a,e) \right] -\mE_{p(x)}\left[\sumE \pi_0(e\,|\,x) \sumA w(x,a) \pi_0(a\,|\,x,e)  q(x,a,e)\right] \notag \\
    & = \mE_{p(x)} \left[ \sumC \pi_0(c\,|\,x)w(x,c) \sumA\sumE \frac{\pi_0(e\,|\,x)\pi_0(a\,|\,x,e)\mathbb{I} \{\phi(x,e) = c\}}{\pi_0(c\,|\,x)} \Delta_{q,\hat{f}}(x,a,e) \right] \notag \\
    & \quad + \mE_{p(x)} \left[ \sumA\sumE \pi_0(a,e\,|\,x) \frac{\pi(a\,|\,x)p(e\,|\,x,a)}{\pi_0(a\,|\,x)p(e\,|\,x,a)} \hat{f}(x,a,e) \right] \notag \\
    & \quad - \mE_{p(x)}\left[\sumE \pi_0(e\,|\,x) \sumA w(x,a) q(x,a,e) \frac{\sumC \pi_0(c\,|\,x) \pi_0(a,e\,|\,x,c) }{\pi_0(e\,|\,x)} \right]  \notag \\
    & = \mE_{p(x)\pi_0(c|x)} \left[ w(x,c)  \sumA\sumE \pi_0(a,e\,|\,x,c) \Delta_{q,\hat{f}}(x,a,e) \right] \notag \\
    & \quad + \mE_{p(x)\pi_0(c|x)} \left[ \sumA\sumE w(x,a)\pi_0(a,e\,|\,x,c) \hat{f}(x,a,e) \right] \notag \\
    & \quad - \mE_{p(x)}\left[ \sumC \pi_0(c\,|\,x) \sumA\sumE w(x,a)\pi_0(a,e\,|\,x,c)  q(x,a,e) \right] \notag \\
    & = \mE_{p(x)\pi_0(c|x)} \left[ \sumA\sumE w(x,a) \pi_0(a,e\,|\,x,c) \sum_{a'\in\calA}\sum_{e'\in\calE} \pi_0(a',e'\,|\,x,c) \Delta_{q,\hat{f}}(x,a',e') \right] \notag \\ 
    & \quad  - \mE_{p(x)\pi_0(c|x)}\left[ \sumA\sumE w(x,a)\pi_0(a,e\,|\,x,c) \Delta_{q,\hat{f}}(x,a,e) \right] \quad \because w(x,c) = \mE_{\pi_0(a,e|x,c)} [w(x,a)] \notag \\
    & = \mE_{p(x)\pi_0(c|x)} \left[ \sumA\sumE w(x,a) \pi_0(a,e\,|\,x,c) \left( \Big( \sum_{a'\in\calA}\sum_{e'\in\calE} \pi_0(a',e'\,|\,x,c) \Delta_{q,\hat{f}}(x,a',e') \Big) - \Delta_{q,\hat{f}}(x,a,e) \right) \right] \label{eq:bias_last_line}
\end{align}
where we used 
\begin{align*}
    w(x,c) 
    & = \frac{\pi(c\,|\,x)}{\pi_0(c\,|\,x)} \\
    & = \frac{1}{\pi_0(c\,|\,x)} \sumA\sumE \pi(a\,|\,x) p(e\,|\,x,a) \mathbb{I}\{\phi(x,e) = c\} \\
    & = \frac{1}{\pi_0(c\,|\,x)} \sumA\sumE \pi(a\,|\,x) \frac{\pi_0(e\,|\,x) \pi_0(a\,|\,x,e)}{\pi_0(a\,|\,x)} \mathbb{I}\{\phi(x,e) = c\} \\
    & = \sumA\sumE \frac{\pi(a\,|\,x)}{\pi_0(a\,|\,x)} \pi_0(a\,|\,x,e) \frac{\pi_0(e\,|\,x)\mathbb{I}\{\phi(x,e) = c\}}{\pi_0(c\,|\,x)} \\
    & = \sumA\sumE w(x,a) \pi_0(a\,|\,x,e) p(e \,|\,x,c) \\
    & = \mE_{\pi_0(a,e|x,c)} [ w(x,a) ] \quad \because \pi_0(a\,|\,x,e) = \pi_0(a\,|\,x,e,c)
\end{align*}

By applying Lemma B.1 of \citet{saito2022off} to Eq.~\eqref{eq:bias_last_line}, we obtain the following expression of the bias.
\begin{align}
    &\biasoffcem \notag \\
    &= \mE_{p(x)\pi_0(c|x)} 
    \Bigg[ 
        \sum_{(a,e)\neq(a',e')} \!\!\!  
        \pi_0(a,e\,|\,x,c) \pi_0(a',e'\,|\,x,c) 
        (\Delta_{q,\hat{f}}(x,a,e) - \Delta_{q,\hat{f}}(x,a',e')) 
        (w(x,a') - w(x,a)) 
    \Bigg], \label{eq:most_general_bias}
\end{align}
where we use $ w(x,a) = \pi(a\,|\,x)/\pi_0(a\,|\,x) = \pi(a,e\,|\,x)/\pi_0(a,e\,|\,x) $. 

Note that if we simplify the problem setting as in the main text, i.e., $\phi: \calX \times \calA \rightarrow \calC$, Eq.~\eqref{eq:most_general_bias} is reduced to
\begin{align*}
    &\biasoffcem \notag \\
    &= \mE_{p(x)\pi_0(c|x)} 
    \Bigg[ 
        \sum_{a<b} 
        \pi_0(a\,|\,x,c) \pi_0(b\,|\,x,c) 
        (\Delta q(x,a,b) - \Delta \hat{f}(x,a,b)) 
        (w(x,b) - w(x,a)) 
    \Bigg] \\
    &= \mE_{p(x)\pi_0(c|x)} 
    \Bigg[ 
        \sum_{a<b} 
        \frac{\pi_0(a\,|\,x)\mathbb{I}\{\phi(x,a)=c\}}{\pi_0(c\,|\,x)} \frac{\pi_0(b\,|\,x)\mathbb{I}\{\phi(x,b)=c\}}{\pi_0(c\,|\,x)}
        (w(x,b) - w(x,a)) 
        (\Delta q(x,a,b) - \Delta \hat{f}(x,a,b)) 
    \Bigg] \\
    &= \mE_{p(x)\pi_0(c|x)} 
    \Bigg[ 
        w(x,c) \! \sum_{a<b:\phi(x,a)=\phi(x,b)=c}  \!
        \pi_0(a\,|\,x,c) \pi_0(b\,|\,x,c) 
        \left( \frac{\pi(b\,|\,x,c)}{\pi_0(b\,|\,x,c)} - \frac{\pi(a\,|\,x,c)}{\pi_0(a\,|\,x,c)} \right)
        (\Delta q(x,a,b) - \Delta \hat{f}(x,a,b)) 
    \Bigg] \\
    &= \mE_{p(x)\pi(c|x)} 
    \Bigg[ 
        \sum_{a<b:\phi(x,a)=\phi(x,b)=c} 
        \pi_0(a\,|\,x,c) \pi_0(b\,|\,x,c) 
        \left( \frac{\pi(b\,|\,x,c)}{\pi_0(b\,|\,x,c)} - \frac{\pi(a\,|\,x,c)}{\pi_0(a\,|\,x,c)} \right)
        (\Delta q(x,a,b) - \Delta \hat{f}(x,a,b)) 
    \Bigg],
\end{align*}
where we use $w(x,a) = \frac{\pi(a | x)}{\pi_{0}(a | x)} = \frac{\pi(a | x, c) \pi(c | x)}{\pi_{0}\left(a | x, c\right) \pi_{0}\left(c | x\right)} = \frac{\pi(a| x, c)}{\pi_{0}(a| x, c)} w(x, c)$ for $\phi(x,a)=c$. 

Theorem~\ref{thm:bias} implies that OffCEM is unbiased under Assumption~\ref{assumption:local_correctness}, as $ \Delta q(x,a,b) - \Delta \hat{f}(x,a,b) =0$ for all $x,a,b$ with $\phi(x,a) = \phi(x,b)=c$ for a given clustering function $\phi(\cdot,\cdot)$.
\end{proof}

\subsection{Proof of Proposition~\ref{prop:variance}} \label{app:variance}
\begin{proof}
We apply the law of total variance several times to obtain the variance of OffCEM. 
\begin{align}
    & \mV_{p(x)\pi_0(a|x)p(r|x,a,e_a)} \left[ w(x,\phi(x,e_a)) (r - \hat{f}(x,a,e_a)) + \hat{f}(x,\pi) \right] \notag \\
    & = \mE_{p(x)\pi_0(a|x)} \left[ \mV_{p(r|x,a,e_a)} \left[ w(x,\phi(x,e_a)) (r - \hat{f}(x,a,e_a)) + \hat{f}(x,\pi) \right] \right] \notag \\
    & \quad + \mV_{p(x)\pi_0(a|x)} \left[ \mE_{p(r|x,a,e_a)} \left[ w(x,\phi(x,e_a)) (r - \hat{f}(x,a,e_a)) + \hat{f}(x,\pi) \right] \right] \notag \\
    & = \mE_{p(x)\pi_0(a|x)} \left[ w(x,\phi(x,e_a))^2 \sigma^2(x,a,e_a) \right] + \mV_{p(x)\pi_0(a|x)} \left[ w(x,\phi(x,e_a)) (q(x,a,e_a) - \hat{f}(x,a,e_a)) + \hat{f}(x,\pi) \right] \notag \\
    & = \mE_{p(x)\pi_0(a|x)} \left[ w(x,\phi(x,e_a))^2 \sigma^2(x,a,e_a) \right] + \mV_{p(x)\pi_0(a|x)} \left[ w(x,\phi(x,e_a)) \Delta_{q,\hat{f}}(x,a) + \hat{f}(x,\pi) \right] \notag \\
    & = \mE_{p(x)\pi_0(a|x)} \left[ w(x,\phi(x,e_a))^2 \sigma^2(x,a,e_a) \right] + \mE_{p(x)} \left[ \mV_{\pi_0(a|x)} \left[ w(x,\phi(x,e_a)) \Delta_{q,\hat{f}}(x,a) + \hat{f}(x,\pi) \right] \right] \notag \\ 
    & \quad + \mV_{p(x)} \left[ \mE_{\pi_0(a|x)} \left[ w(x,\phi(x,e_a)) g(x,\phi(x,e_a)) + \hat{f}(x,\pi) \right] \right] \quad \because \text{Assumption~\ref{assumption:general_local_correctness}}  \notag \\
    & = \mE_{p(x)\pi_0(a|x)} \left[ w(x,\phi(x,e_a))^2 \sigma^2(x,a,e_a) \right] + \mE_{p(x)} \left[ \mV_{\pi_0(a|x)} \left[ w(x,\phi(x,e_a)) \Delta_{q,\hat{f}}(x,a) \right] \right] + \mV_{p(x)} \left[ \mE_{\pi(a|x)} [q(x,a,e_a)] \right] \label{eq:most_general_variance}
\end{align}
where $ \Delta_{q,\hat{f}}(x,a,e) = q(x,a,e) - \hat{f}(x,a,e)$ and $\mE_{\pi_0(a|x)} [ w(x,\phi(x,e_a)) g(x,\phi(x,e_a)) + \hat{f}(x,\pi)]$ = $\mE_{\pi(a|x)} [q(x,a,e_a)]$ as in Section~\ref{app:unbias} under Assumption~\ref{assumption:general_local_correctness}.

If we simplify the problem setting as in the main text, i.e., $\phi: \calX \times \calA \rightarrow \calC$, Eq.~\eqref{eq:most_general_variance} is reduced to the following.
\begin{align*}
    & \mV_{p(x)\pi_0(a|x)p(r|x,a)} \left[w(x,\phi(x,a)) (r - \hat{f}(x,a)) + \hat{f}(x,\pi) \right] \\
    & = \mE_{p(x)\pi_0(a|x)} \left[ w(x,\phi(x,a))^2 \sigma^2(x,a) \right] 
       + \mE_{p(x)} \left[ \mV_{\pi_0(a|x)} \left[ w(x,\phi(x,a)) \Delta_{q,\hat{f}}(x,a) \right] \right] 
       + \mV_{p(x)} \left[ \mE_{\pi(a|x)} [q(x,a)] \right].
\end{align*}
\end{proof}

\section{Detailed Experiment Settings and Results} \label{app:additional_results}

\subsection{Baseline Estimators} \label{app:baselines}
Below, we define and describe the baseline estimators compared in our experiments in detail.

\paragraph{Direct Method (DM).}
DM is defined as follows.
\begin{align*}
    \dm := \frac{1}{n} \sum_{i=1}^n \hat{q}(x_i,\pi) = \frac{1}{n} \sum_{i=1}^n \sumA \pi(a\,|\,x_i) \hat{q} (x_i, a), 
\end{align*}
where $\hat{q}(x,a)$ estimates $q(x,a)$ based on logged bandit data. The accuracy of DM depends on the quality of $\hat{q}(x,a)$. If $\hat{q}(x,a)$ is accurate, so is DM.  However, if $\hat{q}(x,a)$ fails to estimate the expected reward accurately, the final estimator is no longer consistent. As discussed in Appendix~\ref{app:related}, the misspecification issue is challenging, as it cannot be easily detected from observed data~\citep{farajtabar2018more,voloshin2019empirical}. This is why DM is often described as a high bias estimator.

\paragraph{Inverse Propensity Score (IPS).}
IPS is defined as follows.
\begin{align*}
    \ips := \meanN w(x_i, a_i) r_i
\end{align*}
where $w(x,a) := \pi(a\,|\,x) / \pi_0(a\,|\,x)$ is the vanilla importance weight. IPS estimates the value of the target policy by simply re-weighting the observed rewards and is unbiased under common support (Assumption~\ref{assumption:common_support}). Its critical issue, however, is that the variance can be excessive in large action spaces due to the fact that the importance weight inflates. 

\paragraph{Doubly Robust (DR)~\citep{dudik2014doubly}.}
DR is defined as follows.
\begin{align*}
    \dr := \frac{1}{n} \sum_{i=1}^n \left\{ w(x_i,a_i)  (r_i-\hat{q}(x_i, a_i) ) + \hat{q}(x_i,\pi) \right\},
\end{align*}
which combines DM and IPS in a way that reduces the variance. More specifically, DR utilizes the estimated reward function $\hat{q}$ as a control variate. If the expected reward is correctly specified, DR is \textit{semiparametric efficient} meaning that it achieves the minimum possible asymptotic variance among regular estimators~\citep{narita2019efficient}. A problem is that, if the expected reward is misspecified, this estimator can have a larger asymptotic MSE compared to IPS. Moreover, as shown in Eq.~\eqref{eq:dr_variance} and empirically verified by~\citet{saito2022off}, DR can have a very large variance in the presence of many actions. 

\paragraph{Marginalized Inverse Propensity Score (MIPS)~\citep{saito2022off}.}
MIPS is defined as follows.
\begin{align*}
    \mips := \meanN w(x_i, e_i) r_i,
\end{align*}
where $w(x,e) := \pi(e\,|\,x) / \pi_0(e\,|\,x)$ is the marginal importance weight defined based on the marginal embedding distribution induced by the target and logging policies. MIPS is unbiased under common embedding support (Assumption~\ref{assumption:common_embed_support}) and no direct effect (Assumption~\ref{assumption:no_direct_effect}). It was also shown by~\citet{saito2022off} that MIPS provides the following variance reduction in comparison to IPS:
\begin{align}
    n \left( \mV_{\calD} [\ips ] - \mV_{\calD} [\mips ] \right) = \mE_{p(x)\pi_0(e|x)} \left[ \mE_{p(r|x,e)} \left[ r^2 \right] \mV_{\pi_0(a|x,e)} \left[ w(x,a) \right]  \right], \label{eq:mips_variance_reduction}
\end{align}
which is always non-negative and can be substantial when the variance of the vanilla importance weights is large, which is likely for large action spaces. However, it is still possible that the variance of MIPS can be extremely large when action embeddings are high-dimensional and fine-grained. Specifically, the variance of MIPS (under no direct effect from Assumption~\ref{assumption:no_direct_effect}) is given as 
\begin{align*}
    n \mV_{\calD} \big[  \mips \big] 
    & =  \mE_{p(x)\pi_0(e|x)} [ w(x,e)^2 \sigma^2 (x,a) ] + \mE_{p(x)} \left[  \mV_{\pi_0(e|x)} [ w(x,e) \Delta_{q,\hat{q}} (x,e) ] \right]  + \mV_{p(x)} \left[  \mE_{\pi(e|x)} [ q (x,e) ] \right],
\end{align*}
which may become almost identical to the variance of IPS when $w(x,e) \approx w(x,a)$ due to high-dimensional and fine-grained embeddings. To reduce the variance of MIPS, \citet{saito2022off} describe a heuristic procedure to perform action feature selection based on logged data, but this produces bias by violating the no direct effect assumption. Thus, the critical bias-variance dilemma still remains in this estimator, particularly in the presence of high-dimensional action embeddings.

\paragraph{Definitions of the estimators compared in Figure~\ref{fig:ablation} of Section~\ref{sec:synthetic}.}
The following defines the estimators compared in Figure~\ref{fig:ablation} (ablation study) of Section~\ref{sec:synthetic}.
\begin{align*}
    \hat{V}_{\mathrm{+clustering}} (\pi; \calD) & := \meanN w(x_i,\phi(x_i,a_i)) r_i, \\
    \hat{V}_{\mathrm{+regression\,model}} (\pi; \calD) & := \meanN \left\{ w(x_i,e_i) (r_i - \hat{q}(x_i,a_i)) + \hat{q}(x_i,\pi) \right\}, \\
    \hat{V}_{\mathrm{clustering+1step\,reg}} (\pi; \calD) & := \meanN \left\{ w(x_i,\phi(x_i,a_i)) (r_i - \hat{q}(x_i,a_i)) + \hat{q}(x_i,\pi) \right\}, \\
    \hat{V}_{\mathrm{clustering+2step\,reg}} (\pi; \calD) & := \meanN \left\{ w(x_i,\phi(x_i,a_i)) (r_i - \hat{f}(x_i,a_i)) + \hat{f}(x_i,\pi) \right\}, 
\end{align*}
where $\hat{f}(x,a) = \hat{g}_{\psi}(x,\phi(x,a)) + \hat{h}_{\theta}(x,a)$ is obtained by performing the two-step regression procedure described in Section~\ref{sec:two-step_regression} while $\hat{q}(x,a)$ is obtained by simply performing a one-step reward regression as in Eq.~\eqref{eq:one_step_reg}.

$\hat{V}_{\mathrm{+clustering}}$ uses cluster importance weighting instead of vanilla and marginal importance weighting and thus reduces the variance from IPS and MIPS. However, it may produce a large bias as it does not deal with the residual effect at all. $\hat{V}_{\mathrm{+regression\,model}}$ uses a regression model $\hat{q}$ and somewhat reduces the variance of MIPS, but it may still produce a very large variance as it depends on the marginal importance weight. $\hat{V}_{\mathrm{clustering+1step\,reg}}$ is a simple version of our proposed estimator combining cluster importance weighting and a regression model $\hat{q}(x,a)$, but it does not perform the two-step regression procedure and thus may not be optimal in terms of bias. In contrast, $\hat{V}_{\mathrm{clustering+2step\,reg}}$ fully leverages the reward function decomposition of Eq.~\eqref{eq:cem} and thus is the ideal version of our proposed estimator.

\subsection{Additional Synthetic Experiment Setup and Results} \label{app:synthetic}

\subsubsection{Additional Experiment Setup}
This section describes how we defined the synthetic reward function in detail. Recall that, in our synthetic experiments, we synthesized the expected reward function as
\begin{align*}
    q(x,a) = g(x,c_a ) + h_{c_a}(x,a), 
\end{align*}
Specifically, we used the following functions as $g(\cdot,\cdot)$ (cluster effect) and $h_{\cdot}(\cdot,\cdot)$ (residual effect), respectively. 
\begin{align*}
    g(x,c_a ) 
    &= g_{base}(x,\text{one\_hot}_{c_a} ) + u_1 \mathbb{I}\{ (\sum_{d=1}^3 x_d) < 1.5 \} \\
    & \quad + u_2 \mathbb{I}\{ (\sum_{d=3}^8 x_d) < -0.5 \} + u_3  \mathbb{I}\{ (\sum_{d=2}^3 x_d) > 3.0 \} + u_4 \mathbb{I}\{ (\sum_{d=5}^{10} x_d) < 1.0 \}, \\
    h_{c_a}(x,a) &= x^{\top} M_{c_a} \text{one\_hot}_{a} + \theta_{x,{c_a}}^{\top} x +  \theta_{a,{c_a}}^{\top} \text{one\_hot}_{a},
\end{align*}
where $\text{one\_hot}_c$ is the one-hot encoding of the action cluster $c$ and $x_d$ is the $d$-th dimension of the context vector $x$. We use \textbf{obp.dataset.polynomial\_reward\_function} as $g_{base}(\cdot,\cdot)$ and $u_1,\ldots,u_4$ are sampled from a uniform distribution with range $[-3,3]$. $M_{c_a}$, $\theta_{x,{c_a}}$, and $\theta_{a,{c_a}}$ are parameter matrices or vectors sampled from a uniform distribution with range $[-1,1]$ separately for each given action cluster $c_a$.

\subsubsection{Additional Results}
This section reports and discusses additional synthetic experiment results regarding varying levels of noise on the rewards, varying target policies, and varying numbers of action clusters. In particular, we demonstrate that OffCEM works particularly better than other baselines when the reward is noisy, and the target and logging policies differ greatly. We also observe that OffCEM is appealing in particular when there is a fewer number of action clusters. The following also reports and discusses the bias-variance decomposition of the results shown in the main text, providing additional insights.

\paragraph{How does OffCEM perform with varying noise levels, target policies, numbers of action embedding dimensions, and numbers of clusters?}
First, Figure~\ref{fig:varying_noise} evaluates how the level of noise on the rewards affects the comparison of the estimators. To this end, we vary the noise level $\sigma \in \{1.0,2.0,3.0,4.0,5.0\}$, where $\sigma$ is the standard deviation of Gaussian reward noise, i.e., $r \sim \mathcal{N} (q(x, a), \sigma^2)$. We can see from the figures that the variance of IPS, DR, and MIPS grows with increased noise as expected, and OffCEM performs increasingly better than the baselines with larger noise levels. Note that DM is relatively more robust to the increased noise compared to IPS, DR, and MIPS, but it is highly biased in all situations, and our OffCEM performs much better than DM in a range of reward noise settings.

Second, Figure~\ref{fig:varying_eps} compares the estimators with a range of target policies ($\epsilon \in \{0.0, 0.2, 0.4, 0.6, 0.8, 1.0\}$ in Eq.~\eqref{eq:synthetic_target}). Note that a larger value of $\epsilon$ introduces a larger entropy for the target policy, making it closer to the logging policy given that we use $\beta=-0.1$ as default (an extreme case with $\epsilon = 1.0$ produces a uniform random target policy). Figure~\ref{fig:varying_eps} shows that all estimators perform worse for smaller $\epsilon$ as expected, where OPE becomes harder in general. IPS, DR, and MIPS perform worse as their variance increases with decreasing $\epsilon$, while DM performs worse as it produces a larger bias. The variance of OffCEM also increases ever so slightly with decreasing $\epsilon$, but it is often much smaller than those of the baselines. In particular, our estimator becomes increasingly effective against the baseline estimators given a near-deterministic target policy (small $\epsilon$), which is highly prevalent in real-world scenarios given that it is often easier-to-implement than stochastic policies and that the optimal policy is always deterministic (except for the case with non-unique optimal actions).

Finally, Figure~\ref{fig:varying_n_clusters} shows the estimators’ performance when we vary the number of action clusters from 10 to 200. We can see from the figure that the variance of OffCEM becomes larger with increasing number of clusters, but it is generally smaller than those of IPS, DR, and MIPS. When there are many action clusters underlying the data-generating process (e.g., $|\calC|=100,200$), however, we observe that the improvement provided by our estimator becomes slightly smaller due to increased bias. Nonetheless, there is no reason to avoid using OffCEM because the baseline estimators do not outperform ours in any of the experiment settings.

\paragraph{Discussion on the bias-variance decomposition of the synthetic results.}
Next, Figures~\ref{fig:varying_n_val} to~\ref{fig:varying_n_def_actions_ablation} report and discuss the bias-variance decomposition of the synthetic results reported in the main text. First, we can see in Figure~\ref{fig:varying_n_val} that the bias of IPS of MIPS are very small as expected while their variance become extremely large especially for small sample sizes. We can also see that DR provides some reduction in MSE due to reduced variance over IPS and MIPS, however, the variance of our estimator is even smaller and is similar to that of DM while its bias is similar to that of DR, which is very small. We can also confirm from the figure that DM remains highly biased even with increased sample sizes. Figure~\ref{fig:varying_n_val_ablation} reports the bias-variance decomposition of the ablation study with varying sample sizes. In particular, it suggests that using only cluster importance weighting (\textbf{``+clustering"}) reduces the variance of MIPS while it produces a large bias as it ignores the residual effect. In contrast, using only a regression model provides some variance reduction while not producing a large bias, but there is still much room for improvement in terms of variance due to its use of marginal importance weighting. Simply combining these techniques (\textbf{``+clustering and one-step reg"}) is effective enough to outperform MIPS in a rage of sample sizes. However, we also observe that performing two-step regression is beneficial in further reducing the bias. Next, Figure~\ref{fig:varying_n_actions} demonstrates that the variance of IPS, DR, and MIPS become larger for larger numbers of actions while that of OffCEM is much more robust, contributing to its greatly reduced MSEs in large action spaces. In Figure~\ref{fig:varying_n_def_actions}, we observe that the bias of IPS, DR, and MIPS become larger when there exists a larger number of unsupported actions as expected while the bias of OffCEM remains very small by separately accounting for the cluster and residual effects. More specifically, it can unbiasedly estimate the cluster effect and dealing with the residual effect to some extent via the model-based estimation. Thus, the small bias of OffCEM is the main reason for its much better MSE than those of IPS, DR, and MIPS when there are many unsupported actions. Note that the variance of IPS and DR decrease with increased unsupported actions, because the deficiency between logging and target policies become relatively small in that setup. Note also that we can clearly see the benefit of combining the two key techniques (cluster importance weighting and regression model) and that of using two-step regression under varying numbers of actions and numbers of unsupported actions in Figures~\ref{fig:varying_actions_ablation} and~\ref{fig:varying_n_def_actions_ablation}.

\subsection{Additional Real-World Experiment Setup} \label{app:real}

Following previous studies~\citep{farajtabar2018more,dudik2014doubly,wang2017optimal,kallus2020optimal,su2020doubly,saito2021evaluating}, we transform the extreme classification datasets (namely EUR-Lex 4K and Wiki10-31K) to contextual bandit feedback data. In a classification dataset $\{(x_i, a_i)\}_{i=1}^{n}$, we have some feature vector $x_i \in \calX$ and ground-truth label $a_i \in \calA$, which will be considered an action. Since the original datasets have very high-dimensional bag-of-words features, we apply feature dimension reduction based on PCA implemented in scikit-learn~\citep{pedregosa2011scikit}. We also drop the labels/actions that have less than 1 (EUR-Lex 4K) or 9 (Wiki10-31K) positive labels during dataset pre-processing. 

We consider stochastic binary rewards where we first define the base reward function as
\begin{align}
    \label{eq:real_reward} 
    \tilde{q}(x, a)=
    \left\{\begin{array}{ll}1-\eta_a & \text {if $a$ is a positive label} \\ \eta_a - 1 & \text {otherwise}\end{array}\right.
\end{align}
where $\eta_a$ is a noise parameter sampled separately for each action $a$ from a uniform distribution with range $[0,0.2]$. Then, for each data, we sample the reward from a Bernoulli distribution with mean $q(x,a) = \sigma(\tilde{q}(x,a))$ where $\sigma(z) = (1 + \exp(-z))^{-1}$ is the sigmoid function. 

We define the logging policy $\pi_0$ by applying the softmax function to an estimated reward function $\hat{q}(x,a)$ as
\begin{align}
    \pi_0(a \,|\, x) =  \frac{\exp( \beta \cdot \hat{q}(x,a))}{ \sum_{a' \in \calA} \exp( \beta \cdot \hat{q}(x,a')) },
    \label{eq:real_logging}
\end{align}
where we use $\beta=30$ for both datasets. We obtain $\hat{q}(x,a)$ by performing a ridge regression with full-information labels. Note that we use the test data recorded in the original datasets for obtaining a logging policy while we use the training data for performing OPE to make them independent.  We also define the target policy $\pi$ as follows.
\begin{align*}
    \pi(a \,|\, x) = (1 - \epsilon) \cdot \mathbb{I} \big\{a = \argmax_{a' \in \calA} q(x,a') \big\} + \frac{\epsilon}{|\calA|}, 
\end{align*}
where we set $\epsilon=0.05$ in the real-world experiment.

To summarize, we first define the expected reward $q(x,a)$ as in Eq.~\eqref{eq:real_reward} on the training dataset. We then sample discrete action $a$ from $\pi_0$ based on Eq.~\eqref{eq:real_logging}. Finally, we sample the reward from a normal distribution with mean $q(x,a)$ and standard deviation $\sigma=1$. Iterating this $n$ times generates logged bandit data $\calD$ with $n$ independent copies of $(x,a,r)$.

\end{document}